\documentclass{article}
\usepackage[preprint]{corl_2025} %

\usepackage{xcolor}
\definecolor{bblue}{HTML}{002676}  
\definecolor{bgold}{HTML}{FDB515}

\usepackage{booktabs}
\usepackage{pifont}
\newcommand{\xmark}{\ding{55}}  %
\usepackage{xspace}
\newcommand{\algabbr}{R2R2R\xspace}
\newcommand{\algname}{Real2Render2Real\xspace}
\usepackage{makecell}
\usepackage{adjustbox}

\usepackage{amsmath}

\usepackage{microtype}
\usepackage{graphicx}

\usepackage{hyperref}
\usepackage{mathtools}
\usepackage{algorithm}
\usepackage{tikz}
\usetikzlibrary{arrows,automata,positioning}
\usepackage{multirow} 
\usepackage{xspace}

\usepackage{comment}
\usepackage{amssymb} %

\usepackage{dsfont}
\usepackage{tabulary}
\usepackage{tabularx}

\usepackage{wrapfig}

\usepackage{dsfont}
\usepackage{tabulary}

\usepackage{wrapfig}
\usepackage{subcaption}

\usepackage{mathtools}
\usepackage{amsthm}
\usepackage{array}

\usepackage{etoc}

\usepackage[capitalize,noabbrev]{cleveref}

\theoremstyle{plain}

\theoremstyle{definition}

\theoremstyle{remark}

\definecolor{orange}{rgb}{1,0.5,0}
\definecolor{lightsalmonpink}{rgb}{1.0, 0.6, 0.6}
\definecolor{verylightsalmonpink}{rgb}{0.966, 0.805, 0.797}
\definecolor{lightblue}{rgb}{0.862, 0.906, 0.984}
\definecolor{lightyellow}{rgb}{1.0, 0.945, 0.797}
\definecolor{lightgreen}{rgb}{0.835, 0.91, 0.828}
\definecolor{lightpurple}{rgb}{0.879, 0.832, 0.902}

\newcommand{\fsize}{small}
\usepackage[font=\fsize,labelfont=bf,skip=5pt]{caption}
\usepackage{wrapfig}

\usepackage{ragged2e}

\usepackage{listings}

\usepackage{xcolor}
\definecolor{codegreen}{rgb}{0,0.6,0}
\definecolor{codegray}{rgb}{0.5,0.5,0.5}
\definecolor{codepurple}{rgb}{0.58,0,0.82}
\definecolor{backcolour}{rgb}{0.95,0.95,0.92}
\definecolor{coral}{HTML}{FF7F50} 
\lstdefinestyle{mystyle}{
    backgroundcolor=\color{backcolour},   
    commentstyle=\color{codegreen},
    keywordstyle=\color{magenta},
    numberstyle=\tiny\color{codegray},
    stringstyle=\color{codepurple},
    basicstyle=\ttfamily\footnotesize,
    breakatwhitespace=false,         
    breaklines=true,                 
    captionpos=b,                    
    keepspaces=true,                 
    numbers=left,                    
    numbersep=5pt,                  
    showspaces=false,                
    showstringspaces=false,
    showtabs=false,                  
    tabsize=2
}

\usepackage{url}

\hypersetup{
    colorlinks=true,
    linkcolor=black,
    citecolor=black,
    filecolor=cyan,
    urlcolor=magenta
}



\def\figTeaser#1{
    \begin{figure}[#1]
    \centering
    \includegraphics[width=0.99\linewidth]{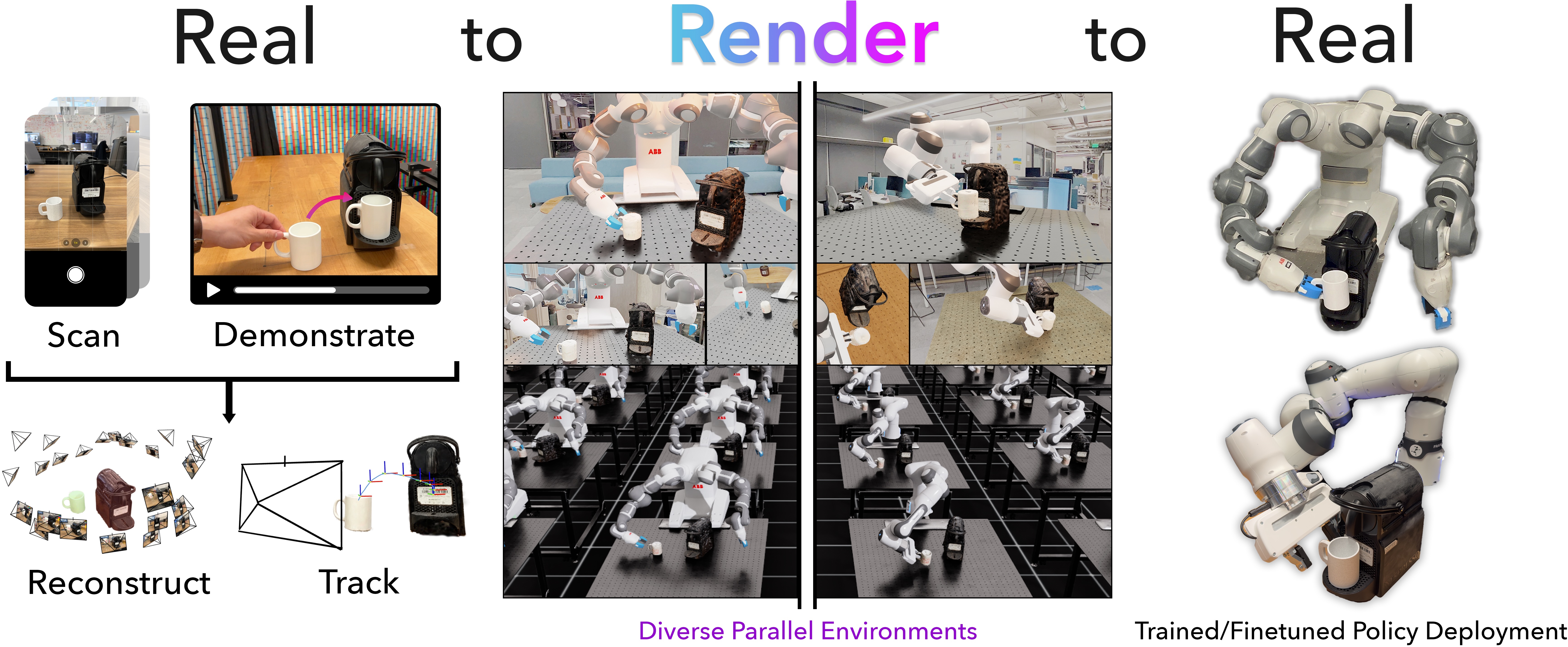}
    \caption{\textbf{Real2Render2Real} generating robot training data for the task of “Put the Mug on the Coffee Maker”. \algabbr takes as input a multi-view object scan and a monocular human demonstration video. R2R2R then synthesizes diverse, domain-randomized robot executions through parallel rendering and outputs paired image-action data for policy training. This pipeline enables scalable learning across tasks and embodiments without teleoperation or object dynamics simulation.}
    \vspace{-2em}
    \label{fig:teaser}
    \end{figure}
}

\def\figScaleTime#1{
    \begin{figure}[#1]
    \centering
        \includegraphics[width=0.99\linewidth]{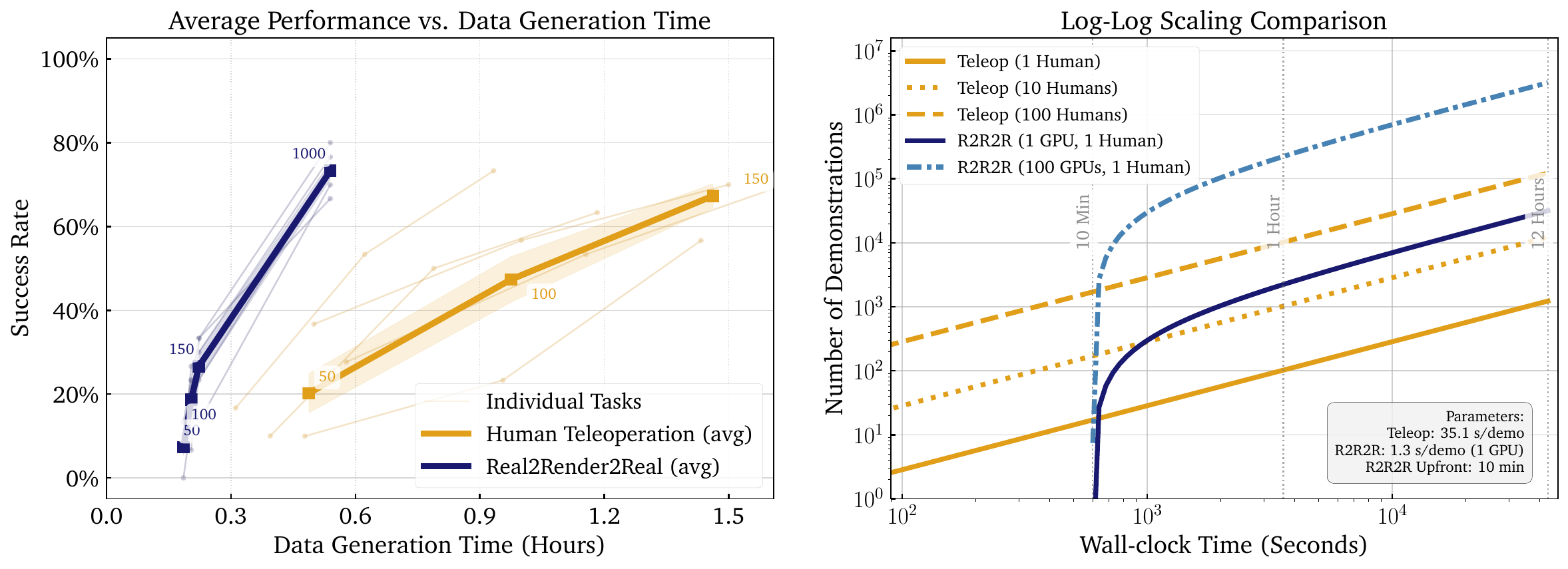}
    \caption{
    \textbf{Data Generation Efficiency and Average Policy Performance Across Manipulation Tasks.} \textbf{(Left)} Performance visualization displaying both task-specific outcomes (faint background lines) and cross-task averages (bold lines with error shading) for policies trained on real (1 human teleoperator) vs. synthetic data (1 human, 1 GPU). The points labeled by demonstration count (50-1000) highlight the scaling in performance and R2R2R's significant throughput advantage, with individual task trajectories illustrating the variance across different manipulation scenarios. \textbf{(Right)} Log-log scale comparison showing data generation throughput between R2R2R (1-100 GPUs) and human teleoperation (1-100 operators) over a 12-hour period. \algabbr needs an upfront time of 10 minutes for human to scan the objects, demonstrate the task, reconstruct the objects and track their trajectory, where subsequentially no human is involved. On a single NVIDIA 4090 GPU, on average, trajectories will be generated at ~27x the speed of a single human teleoperator without needing robot hardware. 
    }
    \label{fig:sclaing_plot_time}
    \vspace{-1em}
    \end{figure}
}

\def\figAvgPerf#1{
    \begin{figure}[#1]
    \centering
        \includegraphics[width=\linewidth]{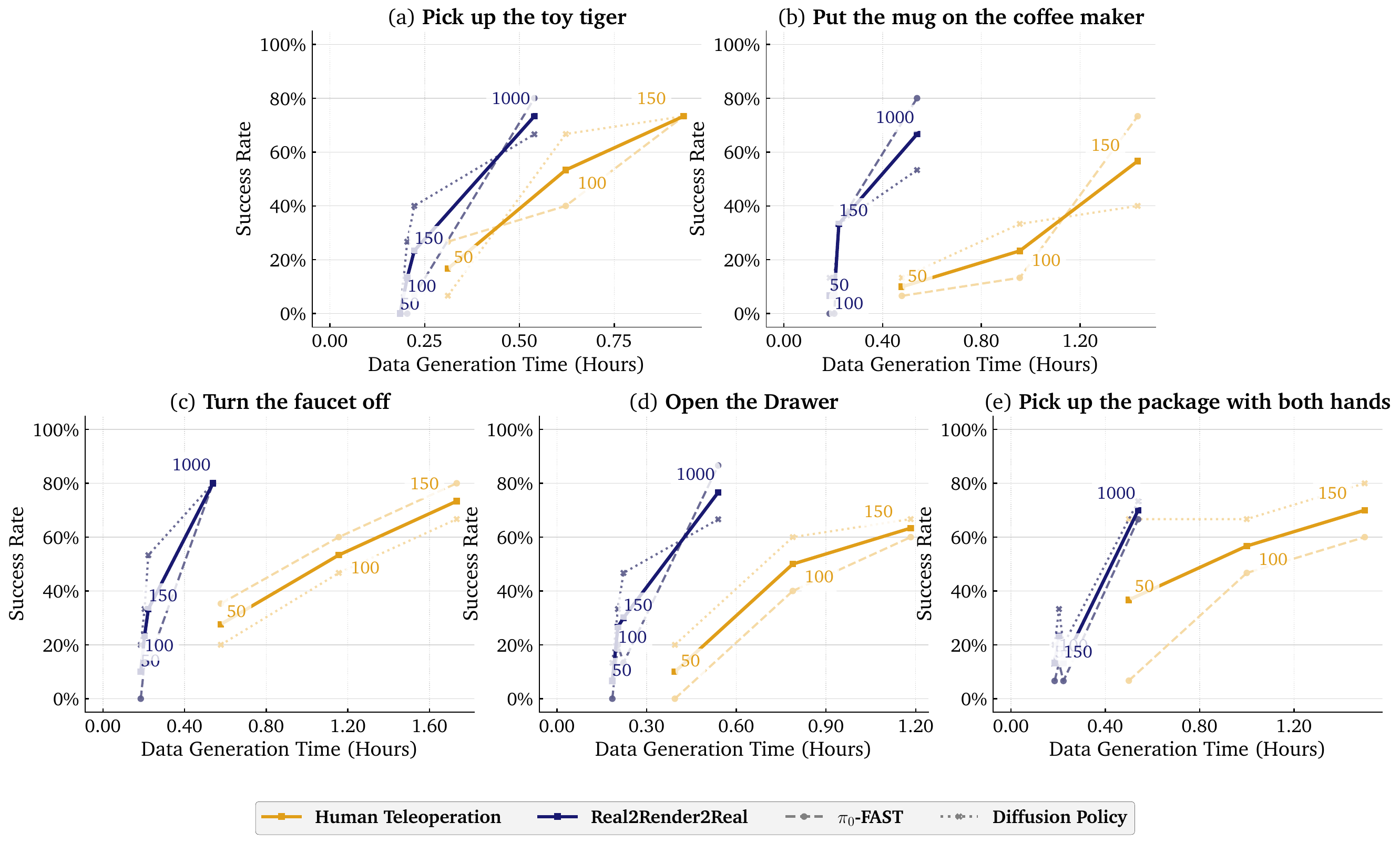}
    \caption{
        \textbf{Physical Experiments Comparing Real2Render2Real to Human Teleoperation Data Efficiency}
        Task success rate is plotted against data generation time in hours.
        Solid lines represent performance averaged across $\pi_0$-FAST and Diffusion Policy.
        The Real2Render2Real line (\textcolor{bblue}{blue} square) includes points corresponding to 50, 100, 150, and 1000 trajectories generated by a single Nvidia RTX 4090.
        The Human Teleoperation line (\textcolor{bgold}{gold} square) includes points corresponding to 50, 100, and 150 trajectories.
        The Real2Render2Real data generation time includes a 10-minute setup cost, while the Human Teleoperation time is based on the real trajectory collection time of 150 demonstrations. Exact numbers for evaluation results can be found in \cref{sec:eval_results}. 
    }
    \label{fig:avg_performance_vs_time}
    \vspace{-1.5em}
    \end{figure}
}

\def\tableRWComp#1{
\begin{table}[#1]
\centering

\begin{adjustbox}{width=\columnwidth,center}
\begin{tabular}{lcccccccc}
\toprule
& \makecell{\textbf{Tele-Op}\\\textbf{Free}} 
& \makecell{\textbf{RL}\\\textbf{Free}}
& \makecell{\textbf{Phys. Engine}\\\textbf{Free}} 
& \makecell{\textbf{Robot}\\\textbf{Agnostic}}
& \makecell{\textbf{One-to-Many}\\ \textbf{Trajectories}}  
& \makecell{\textbf{Articulated}\\ \textbf{Objects}}  \\
\midrule
CASHER~\cite{torne2024robot} & \xmark & \xmark & \xmark  & \checkmark & \checkmark & \checkmark \\ %
RoboVerse~\cite{geng2025roboverse} & \xmark & \xmark & \xmark  & \checkmark & \xmark & \checkmark \\ %
RoboGSim~\cite{li2024robogsim} & \xmark & \checkmark & \checkmark  & \xmark & \xmark  & \xmark\\ %
RoVi-Aug~\cite{chen2024rovi} & \xmark & \checkmark & \checkmark & \checkmark & \xmark & \checkmark\\ %
Video2Policy~\cite{ye2025video2policy} & \checkmark & \xmark  & \xmark  & \checkmark & \checkmark & \checkmark\\ %
MimicGen~\cite{mandlekar2023mimicgen} & \xmark & \checkmark  & \xmark & \checkmark  & \checkmark & \xmark\\ %
DexMimicGen~\cite{jiang2024dexmimicgen} & \xmark & \checkmark  & \xmark & \checkmark & \checkmark & \xmark\\ %
    Phantom~\cite{lepert2025phantom} & \checkmark & \checkmark & \checkmark  & \checkmark & \xmark & \checkmark\\ %
DemoGen~\cite{xue2025demogen} & \xmark & \checkmark & \checkmark & \xmark & \checkmark & \xmark \\ %
AR2-D2~\cite{duan2023ard} & \checkmark & \checkmark & \checkmark & \checkmark & \xmark & \checkmark \\ %
\midrule
\textbf{Real2Render2Real} & \textbf{\checkmark} & \textbf{\checkmark} & \textbf{\checkmark}& \textbf{\checkmark} & \textbf{\checkmark} & \textbf{\checkmark} \\
\bottomrule
\end{tabular}
\end{adjustbox}
\vspace{3.0pt}
\caption{\textbf{Comparison of Robot Data Generation Methods.} \textit{Real2Render2Real} requires no teleoperation, eliminates reliance on reward engineering, reinforcement learning, or accurate asset physics modeling, and provides object-centric demonstrations directly extracted from a video where humans interact with the objects. It also supports various robot embodiments, and generates multiple varied trajectories from a single demonstration.}
\label{tab:comparison}
\vspace{-1.5em}
\end{table}
}

\def\tableSynthDataEval#1{
\begin{table}[htbp]
    \centering
    \begin{tabular}{lcccc}
        \toprule  %
        Num. Trajectories & 50 & 100 & 150 & 1000\\
        \midrule
        & \multicolumn{4}{c}{\textbf{Pick up the tiger}}\\
        \midrule
        Diffusion Policy & 0.0\% & 26.7\% & 40.0\% & 66.6 \% \\
        $\pi_0$-FAST (Finetuned) & 0.0\% & 0.0\% & 6.7\% & 80.0\% \\
        \midrule
        & \multicolumn{4}{c}{\textbf{Put the mug on the coffee maker}}\\
        
        \midrule
        Diffusion Policy & 13.3\% & 13.3\% & 33.3\% & 53.3\% \\
        $\pi_0$-FAST (Finetuned) & 0.0\% & 0.0\% & 33.3\% & 80.0\%\\

        \midrule
        & \multicolumn{4}{c}{\textbf{Pick up the package with both hands}}\\
        \midrule
        Diffusion Policy & 20.0\% & 33.3\% & 20.0\% & 73.3\% \\
        $\pi_0$-FAST (Finetuned) & 6.6\% & 13.3\% & 6.6\% & 66.6 \% \\

        \midrule
        & \multicolumn{4}{c}{\textbf{Open the drawer}}\\
        \midrule
        Diffusion Policy & 13.3\% & 33.3\% & 46.7\% & 66.6 \%\\
        $\pi_0$-FAST (Finetuned) & 0.0\% & 20.0\% & 13.3\% & 86.6\% \\

        \midrule
        & \multicolumn{4}{c}{\textbf{Turn the faucet off}}\\
        \midrule
        Diffusion Policy & 20.0\% & 33.3\% & 53.3\% & 
        80.0\% \\
        $\pi_0$-FAST (Finetuned) & 0.0\% & 13.3\% & 13.3\% &
        80.0\%\\
        
        \bottomrule%
        
    \end{tabular}
    \vspace{5.5pt}
    \caption{\textbf{Physical Evaluation of Policies Trained Exclusively with R2R2R Data.}}
    \label{tab:synth_eval}
\end{table}
}

\def\tableRealDataEval#1{
\begin{table}[htbp]
    \centering
    \begin{tabular}{lccc}
        \toprule  %
        Num. Trajectories & 50 & 100 & 150\\
        \midrule
        & \multicolumn{3}{c}{\textbf{Pick up the tiger}}\\
        \midrule
        Diffusion Policy & 6.67\% & 66.7\% & 73.3\% \\
        $\pi_0$-FAST (Finetuned) & 26.7\% & 40.0\% & 73.3\% \\
        \midrule
        
        & \multicolumn{3}{c}{\textbf{Put the mug on the coffee maker}}\\
        \midrule
        Diffusion Policy & 13.3\% & 33.3\% & 40.0\%  \\
        $\pi_0$-FAST (Finetuned) & 6.6\% & 13.3\% & 73.3\%\\

        \midrule

        & \multicolumn{3}{c}{\textbf{Pick up the package with both hands}}\\
        \midrule
        Diffusion Policy & 66.7\% &  66.7\% & 80.0\% \\
        $\pi_0$-FAST (Finetuned) & 6.7\% & 46.7\% & 60.0\% \\
        
        \midrule
        & \multicolumn{3}{c}{\textbf{Open the drawer}}\\
        \midrule
        Diffusion Policy & 20.0\% & 60.0\% & 66.7\% \\
        $\pi_0$-FAST (Finetuned) & 0.0\% & 40.0\% & 60.0\%\\

        \midrule
        & \multicolumn{3}{c}{\textbf{Turn the faucet off}}\\
        \midrule
        Diffusion Policy & 20.0\% & 46.7\% & 66.7\% \\
        $\pi_0$-FAST (Finetuned) & 35.3\% & 60.0\% & 80.0\%\\        

        \bottomrule%
        
    \end{tabular}
    \vspace{5.5pt}
    \caption{\textbf{Physical Evaluation of Policies Trained Exclusively with Tele-Operated Data.}}

    \label{tab:real_eval}
\end{table}
}

\title{Real2Render2Real: Scaling Robot Data\\Without Dynamics Simulation or Robot Hardware}

\author{
    Justin Yu$^{1,}$\thanks{Equal Contribution}~, ~Max Letian Fu$^{1,*}$, ~Huang Huang$^1$, ~Karim El-Refai$^1$, \\\textbf{~Rares Andrei Ambrus$^2$, ~Richard Cheng$^2$, ~Muhammad Zubair Irshad$^2$, ~Ken Goldberg$^1$}\\
  \vspace{-5pt}
  \\
  $^1$University of California, Berkeley, ~$^2$Toyota Research Institute
}

\begin{document}
\maketitle

\begin{abstract}
Scaling robot learning requires vast and diverse datasets. Yet the prevailing data collection paradigm—human teleoperation—remains costly and constrained by manual effort and physical robot access. 
We introduce \textbf{Real2Render2Real (R2R2R)}, a novel approach for generating robot training data without relying on object dynamics simulation or teleoperation of robot hardware.
The input is a smartphone-captured scan of one or more objects and a single video of a human demonstration. R2R2R renders thousands of high visual fidelity robot-agnostic demonstrations by reconstructing detailed 3D object geometry and appearance, and tracking 6-DoF object motion. R2R2R uses 3D Gaussian Splatting (3DGS) to enable flexible asset generation and trajectory synthesis for both rigid and articulated objects, converting these representations to meshes to maintain compatibility with scalable rendering engines like IsaacLab but with collision modeling off. Robot demonstration data generated by R2R2R integrates directly with models that operate on robot proprioceptive states and image observations, such as vision-language-action models (VLA) and imitation learning policies. 
Physical experiments suggest that models trained on R2R2R data from a single human demonstration can match the performance of models trained on 150 human teleoperation demonstrations.
Project page: \url{https://real2render2real.com}.

\end{abstract}
\keywords{Robot Datasets, Imitation Learning, Data Augmentation}

\section{Introduction}
\begin{quote}
The great power of general purpose methods~\ldots~[is that they] continue to scale with increased computation.

\hfill --- Richard Sutton, \textit{The Bitter Lesson} (2019)
\end{quote}

Robotics has long benefited from computational scalability—methods like probabilistic planning, trajectory optimization, and reinforcement learning have driven significant progress in agile locomotion~\cite{kavraki1996probabilistic, lavalle2001rapidly, tassa2012synthesis, posa2016optimization, mastalli2020crocoddyl, kumar2021rma, radosavovic2024real}. Dexterous manipulation, however, presents unique challenges: it requires fine-grained visual perception tightly coupled with robot control and kinematics to interact with objects and alter the environment. Many systems address this by explicitly separating perception from planning and control, achieving strong performance in structured environments~\cite{lenz2015deep, pinto2016supersizing, mahler2017dex, mahler2019learning}, especially when assumptions about scene geometry, object placement, and sensing modalities hold. Yet such pipelines often rely on task-specific perception modules and carefully controlled environments, limiting flexibility in more unstructured, dynamic, or visually diverse settings.

In the hope of addressing open-world manipulation tasks, inspired by large language models (LLMs) and vision-language models (VLMs)~\cite{openai2023gpt4, geminiteam2024geminifamilyhighlycapable, lu2024qwenvl, liu2023llava}, recent efforts have explored end-to-end generalist robot policies~\cite{black2410pi0, bjorck2025gr00t, team2025gemini, openvla, chi2023diffusion, octo_2023, huang2025otter, zhao2023learning, Ze2024DP3, fu2024icrt}—models that learn directly from raw sensory input and promise capabilities like language instruction following, task transfer, and in-context learning. Yet training such models at scale remains limited by data: the largest human teleoperation datasets are over 100,000× smaller than the corpora used to train frontier LLMs and VLMs~\cite{goldberg2025data, mirchandani2024so}, and are constrained by the cost, speed, and embodiment-specific nature of human teleoperated data collection.

\figTeaser{t!}

Other vision-language subfields have faced similar data scarcity—and overcome it through \emph{computational} data generation. Structure-from-motion, detection, and depth pipelines now routinely produce pseudo-labels to bootstrap large models; for instance, SpatialVLM synthesizes two billion spatial-reasoning QA pairs~\cite{Chen_2024_CVPR}, while RAFT~\cite{teed2020raft}, DUSt3R~\cite{dust3r_cvpr24}, MonST3R~\cite{zhang2024monst3r}, Zero-1-to-3~\cite{liu_zero-1--3_2023}, and MVGD~\cite{guizilini2025zeroshotnovelviewdepth} all rely on pseudo ground-truth derived from multi-view geometry pipelines (e.g., COLMAP~\cite{schoenberger2016sfm}) to supervise dense 3D prediction tasks. These successes suggest an analogous question for robotics:

\emph{Can we computationally scale robot vision-action data -- while not requiring dynamics simulation or human teleoperation -- to train robot learning models?}

Prior efforts have turned to physics-based simulation, where trajectories are synthesized via reinforcement learning or motion planning in virtual environments~\cite{katara2024gen2sim, wang2023robogen, ye2025video2policy}. While modern simulators offer high throughput and support large-scale parallelization, they face several fundamental limitations: many commonly used simulators fail to satisfy basic Lagrangian mechanics, such as conservation of energy or momentum~\cite{geng2025roboverse}; accurately modeling complex object interactions often demands extensive parameter tuning and hand-crafting of contact properties~\cite{torne2024robot}; generating high-quality, compliant, and intersection-free assets for simulation remains labor-intensive, as collision modeling requires careful handling of geometry, friction, and deformation~\cite{li2020incremental, kim2022ipc}. \algabbr avoids these challenges by discarding dynamics: instead of simulating forces or contacts, we directly set object and robot poses per frame using the IsaacLab package \cite{mittal2023orbit} purely as a photorealistic, parallelized rendering engine by setting all objects as kinematic rather than dynamic bodies. This approach respects robot kinematics while avoiding the complexities of contact modeling, naturally aligning with vision-based policies trained from RGB images and proprioceptive inputs.

We introduce \textbf{Real2Render2Real (\algabbr)}, a pipeline for generating large-scale synthetic robot training data from a smartphone object scan and a human demonstration video. \algabbr scales \textit{trajectory diversity} while preserving visual accuracy: it extracts 6-DoF object part trajectories from the video using object pose tracking and generates corresponding robot executions via differential inverse kinematics under randomized object initializations. Starting from a multi-view scan, it reconstructs 3D object geometry and appearance, supports both rigid and articulated objects via part-level decomposition, and uses 3D Gaussian Splatting to produce mesh assets. The resulting trajectories include robot proprioception, end-effector actions, and paired RGB observations rendered under varied lighting, camera pose, and object placement—making them directly compatible with modern imitation learning policies such as vision-language-action models and diffusion models. By eliminating the need for dynamics simulation or robot hardware, \algabbr enables accessible and scalable robot data collection, allowing anyone to contribute by capturing everyday object interactions with a smartphone.

\figScaleTime{t!}

This paper makes three contributions. First, we present \textit{Real2Render2Real (\algabbr)}, a novel framework that synthesizes diverse, physically grounded observation–action pairs using only smartphone-captured videos: a multi-view object scan and a human demonstration video—without requiring dynamics simulation or robot hardware. Second, we demonstrate that this data is compatible with modern vision-language-action (VLA) and imitation learning policies, including both transformer-based and diffusion-based architectures that operate from RGB and proprioceptive input. Third, we show that policies trained on \algabbr-generated data based on one human demonstration can match the performance of those trained on 150 human teleoperation demonstrations, across 1,050 physical robot evaluations, while requiring significantly less time to generate.

\section{Related Work}
\label{sec:related_work}

\textbf{Robot Data Collection Paradigms.}
Scaling robot learning has traditionally relied on two paradigms: data from industrial deployments and data from human teleoperation. Industrial robot logs~\cite{Mitash2023, sohn2024rfm1, satish2025prime1} scale with production throughput but are often task- and embodiment-specific. In contrast, teleoperation datasets~\cite{teed2021droid, 2024openxembodiment, ebert2021bridge, walke2023bridgedata, fang2023rh20t} offer greater visual and task diversity but remain bottlenecked by human effort and real-time collection. At the same time, the rise of generalist robot policies~\cite{black2410pi0,openvla,rt22023arxiv, bjorck2025gr00t, team2025gemini, chi2023diffusion, octo_2023, huang2025otter, zhao2023learning, Ze2024DP3, fu2024icrt}—capable of performing diverse manipulation tasks from raw observations—has amplified the need for scalable, diverse, and high-quality training data. Yet the scale of current robot datasets remains orders of magnitude below that of their vision and language counterparts~\cite{goldberg2025data, mirchandani2024so}.

\textbf{Procedural Robot Data Generation}. To address the challenge of robot data scaling, many works have studied procedural data generation to automate robot data collection for pre-defined tasks. Many works~\cite{levine2018learning,kerr2022self,wilcox2022learning, chen2022efficiently,fu2023safe,radosavovic2023robot} use pre-defined motion primitives, optionally with a perception module, to automate data collection using a real robot, with automatic scene reset. While reducing human interventions, they still require robot hardware for data collection, limiting scalability. More recently, simulation data generation has emerged as a scalable alternative to real-world collection, parallelizing data generation without physical robot hardware. Utilizing the privileged information from the simulator, ~\citet{mahler2017dex} generates large and diverse data for robot grasping. ~\citet{katara2024gen2sim, wang2023robogen} generate large-scale robot data in simulation using reinforcement learning, trajectory optimization, and motion planning. MimicGen~\cite{mandlekar2023mimicgen} synthesizes diverse simulations from a single human tele-operation sequence, combining motion planning and trajectory replaying. Despite efforts to bridge the sim-to-real gap through domain randomization~\cite{tobin2017domain}, improved asset and scene generation~\cite{katara2024gen2sim, wang2023robogen}, the resulting simulation data often exhibit significant visual discrepancies from real-world observations, requiring co-training on real data to enable effective transfer~\cite{maddukuri2025simandreal}. 

\tableRWComp{t!}
\textbf{Real2Synthetic Data Generation}. To mitigate this visual domain gap, some works augment and repurpose real RGB data instead of synthesizing it from scratch. For example, ~\citet{chen2024rovi} employs generative models for inpainting robot embodiment features into real images, enabling data synthesis for robots with different morphologies. However, such approaches still require human teleoperation for initial demonstrations. Further, they lack the ability to generate additional diverse trajectories beyond the provided demonstrations. Similarly, ~\citet{lepert2025phantom, duan2023ard} use hand-pose tracking to guide inpainted robot end-effector trajectories from human demonstrations. While these methods reduce the need for direct teleoperation, they typically generate only a single trajectory per video and lack support for computationally-scaled trajectory diversity. In contrast, \algabbr can generate multiple, diverse robot trajectory renderings and action rollouts from a single human demonstration. Policies trained solely on \algabbr-generated data achieve comparable real-world performance with those trained on human teleoperation data.

\textbf{Real2Sim2Real Data Generation}. 
To generate diverse trajectories from a single demonstration while bridging the sim-to-real gap, many methods follow a Real2Sim2Real paradigm—using real-world observations to build simulated environments for policy learning. Prior work~\cite{lim2022planarrobotcastingreal2sim2real} shows that tuning physics parameters can reduce dynamics mismatch, but large visual domain gaps still necessitate test-time perception modules.
Recent methods \cite{ye2025video2policy, pfaff2025scalable, geng2025roboverse, jiang2024dexmimicgen} reduce this visual gap by constructing digital twins or “digital cousins”\cite{dai2024automated} from real scans. These approaches vary in their reliance on teleoperation, simulation, and trajectory diversity—but many still depend on teleoperated demos, handcrafted rewards, or accurate physics models, limiting scalability. For example, DexMimicGen\cite{jiang2024dexmimicgen} uses fixed simulation assets; RoboVerse~\cite{geng2025roboverse} supports only rigid objects; and RialTo~\cite{torne2024reconciling} and CASHER~\cite{torne2024robot} require manual articulation labeling and reward engineering. While Video2Policy~\cite{ye2025video2policy} avoids reward tuning via vision-language models, it still requires test-time object detection due to visual mismatches.
These pipelines also rely on physics engines, which demand high-fidelity meshes for collision checking and extensive tuning. RoboGSim~\cite{li2024robogsim} avoids simulation but lacks support for trajectory diversity from a single demo.
In contrast, \algabbr addresses these limitations by: (1) extracting object trajectories from human videos,
(2) segmenting object parts automatically,
(3) rendering realistic observations to remove reliance on test-time perception,
(4) eliminating the need for collision modeling and detailed meshes, and
(5) generating diverse trajectories from a single demonstration.

\section{Assumptions}

We assume objects are rigid or articulated, and manipulated on a table-top setup 
under quasistatic conditions. Object surfaces are assumed to exhibit low specularity to support robust geometry reconstruction and visual feature extraction. We also assume that during human demonstrations, objects are not placed in configurations that lead to complete mutual occlusion. 
Approximate camera poses relative to the robot in the physical setup are assumed to be available, enabling the generation of observations from nearby viewpoints during data collection.
Learned policies take RGB image observations and robot's proprioceptive states as inputs.

\section{Method}
\label{sec:method}

Real2Render2Real (\algabbr) is a data generation pipeline for synthesizing diverse robot demonstration data consisting of RGB-action pairs from a single human demonstration and multi-view object scan. \algabbr consists of three primary stages: (1) \textbf{real-to-sim asset and trajectory extraction}, where rigid or articulated object geometry and part trajectories are extracted from real-world smartphone captures; (2) \textbf{augmentation}, where object initialization is randomized and object motion trajectories are interpolated if appropriate; and (3) \textbf{parallelized rendering}, where diverse photorealistic robot executions are generated using IsaacLab~\cite{mittal2023orbit}, scalable with the amount of available GPU memory and the numbers of GPUs.

\subsection{Real-to-Sim Asset Extraction}
We extract 3D object assets from smartphone scans using a two-stage process inspired by \cite{kerr2024robot, yu2025pogs}. First, we reconstruct object geometry and appearance using 3D Gaussian Splatting (3DGS)~\cite{kerbl20233d}, then apply GARField~\cite{garfield2024} to segment the scene into semantically meaningful parts by lifting 2D masks into 3D. This enables both object-level and part-level decomposition, including articulated components. To support mesh-based rendering, the resulting Gaussian groups are converted into textured triangle meshes via an extended version of ~\cite{guedon2023sugar}.

\begin{figure}[t]
    \centering
        \includegraphics[width=0.99\linewidth]{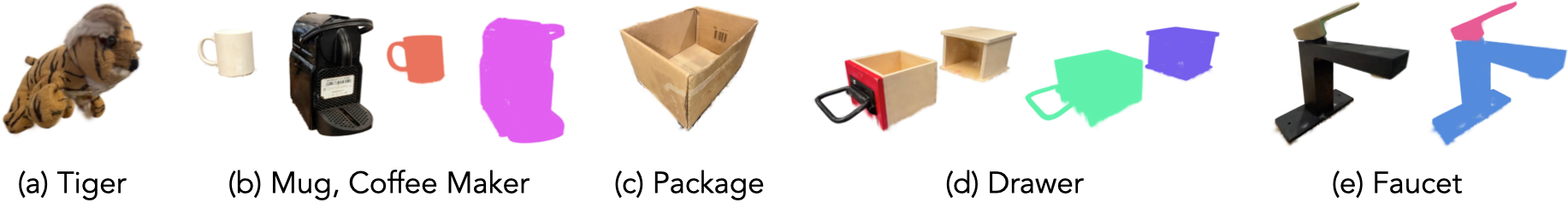}
    \caption{\textbf{3D Gaussian Splat Object Reconstructions} with part-level segmentations derived from feature-based grouping. Objects are reconstructed and segmented into rigid or articulated components using GARField~\cite{garfield2024}.}
    \label{fig:assets_gs}
\end{figure}

\subsection{Real-to-Sim Trajectory Extraction}
\label{ssec:real_sim_traj_extrac}
Given a smartphone video of a human manipulating the scanned objects, \algabbr extracts the 6-DoF part motion of the object and its parts using 4D Differentiable Part Modeling (4D-DPM) introduced in \cite{kerr2024robot}. Each 3DGS object part is embedded with pre-trained DINO features, enabling part pose optimization through differentiable rendering. We extend \cite{kerr2024robot}'s implementation to track single or multiple rigid objects, as well as articulated ones, from demonstration videos.

While there are many alternative pipelines that convert real images into 3D assets, we adopt 3DGS-to-mesh conversion for two key reasons: (1) it enables background–foreground segmentation and part decomposition via 3D grouping~\cite{garfield2024}, which is critical for extracting object part-specific trajectories from monocular human demonstrations; and (2) it maintains compatibility with both 4D-DPM trajectory reconstruction and mesh-based rendering engines, allowing seamless integration into our large-scale rendering pipeline. This process requires no fiducials or hardware beyond a smartphone camera, making it well-suited for scalable and accessible real-to-sim data generation.

\textbf{Interpolation Methods for Object Trajectory Diversity:} A key contribution of \algname is the ability to synthesize multiple valid 6-DoF object trajectories from a single human demonstration. In the case of multiple rigid objects that interact, (e.g. putting a mug on a coffee-maker) the original demonstration is valid only for a specific initial object configuration, and naively replaying it from a new initial pose would fail. To address this, we introduce a suite of trajectory interpolation and resampling techniques that adapt the original trajectory to new start and end poses while preserving its semantic intent.
\begin{wrapfigure}{r}{0.39\linewidth}
    \includegraphics[width=\linewidth]{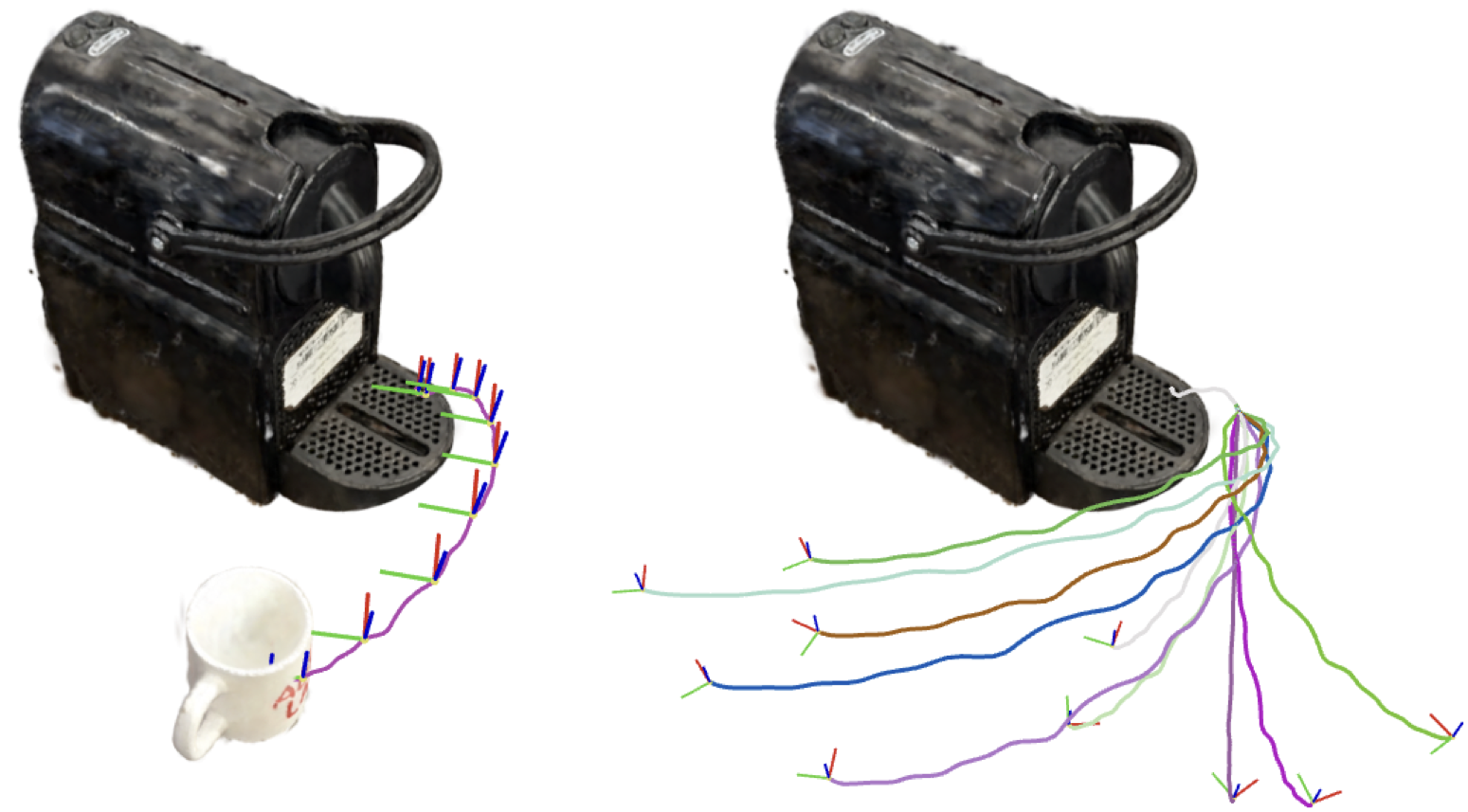}
    \caption{\textbf{Trajectory Interpolation} -- \algabbr adapts object motion to varied start/end configurations via spatial normalization and Slerp.}
    \label{fig:traj_interp}
    \vspace{-10pt}  %
\end{wrapfigure}

We begin with a reference trajectory $\tau \in \mathbb{R}^{T \times 7}$ consisting of $T$ waypoints provided by the part tracking from the demonstration video, each encoding an object orientation (quaternion) and position. Given a new initial pose $\mathbf{x}_{\text{start}}$ and the desired end pose $\mathbf{x}_{\text{end}}$ from human demonstration, we apply a spatial normalization that transforms the original trajectory into a canonical space. We compute the affine transform between the original and target endpoint poses, apply it to the translational component of the trajectory, and interpolate keyframe orientations using spherical linear interpolation (Slerp). This results in a new trajectory $\tau'$ that begins and ends at the desired poses while respecting the structure of the original motion. While these trajectories preserve high-level semantic intent, they are generated without explicit collision avoidance and may result in infeasible paths when initialized behind occluding objects. To mitigate this, we apply a sampling heuristic that biases the distribution of initial placements away from the goal pose (see Fig~\ref{fig:traj_interp}).

\textbf{Grasp Pose Sampling:} R2R2R estimates 3D hand keypoints from the demonstration video using~\cite{pavlakos2024reconstructing}, then determines object-hand interactions by computing the Euclidean distance between keypoints (index fingertip and thumb) and the centroids of all segmented object parts. This produces a distance matrix indexed over time and object parts. We identify the grasped part as the one with the minimum aggregate distance across the trajectory, effectively selecting the object most consistently proximal to the hand throughout the demonstration. To generate physically plausible grasps, we sample 3DGS means to construct a coarser triangle mesh (distinct from the high-resolution rendering mesh), apply surface smoothing and decimation to obtain consistent normals, and use an analytic antipodal grasp sampler following~\cite{mahler2017dex} to determine candidate grasp axes. For bimanual tasks, this process is applied independently per hand to infer separate object associations and grasps, supporting coordinated actions such as lifting or stabilization.

\textbf{Differential Inverse Kinematics:} For each grasp and object trajectory pair, we solve a differential inverse kinematics problem using PyRoki~\cite{pyroki2025}. The solver computes smooth joint-space trajectories that induce the desired object motion across the pre-grasp, grasp, and post-grasp phases. Crucially, our method does \textit{not require modeling object dynamics or simulating physics interactions}. %
Instead of solving for joint torques that would physically induce object movement (as in dynamic simulation), we assume the object rigidly follows the trajectory during contact. This kinematic assumption avoids challenges like contact modeling, compliance, or friction estimation. The solver simply ensures that robot kinematics can track the desired object motion subject to joint position limits, and during pre- and post-grasp phases we additionally include smoothness and velocity limits constraints, which generates valid grasp approach motions. 

\textbf{Rendering Diverse Environment Contexts:} To improve robustness, we apply extensive domain randomization across both scene geometry and rendering parameters. This includes randomized lighting conditions (e.g., intensity, color temperature), camera extrinsics (uniformly sampled up to 2cm translation and 5$^{\circ}$ rotation), and object initial poses (sampled within a workspace-relevant range). By modeling 3D object-centric representations, we can apply these augmentations directly during rendering. Changes in camera pose or lighting do not affect the underlying kinematic rollout, allowing R2R2R to generate diverse visual contexts from a single demonstration. These augmentations expand the data distribution and improve generalization by mitigating the appearance gap and covariate shift between synthetic demonstrations and real-world deployment.

\textbf{High-Throughput Rendering:} IsaacLab~\cite{mittal2023orbit} supports GPU-parallel execution of multiple environment contexts using tile-based rendering, deep-learning super sampling (DLSS), and mesh asset instancing. On a single NVIDIA RTX 4090, R2R2R uses the IsaacLab framework to render complete robot demonstrations at an average rate of 51 demonstrations per minute—compared to 1.7 demonstrations per minute via human teleoperation—yielding over a \textbf{27$\times$} speedup. This throughput scales linearly with the number of rendering GPUs, as depicted in Figure~\ref{fig:sclaing_plot_time}. Data generation/collection time per task can be found in \cref{tab:task_data_coll_time}.

\subsection{Policy Learning}
\vspace{-1em}
 We consider two modern imitation learning architectures: Diffusion Policy~\cite{chi2023diffusion} and $\pi_0$-FAST~\cite{pertsch2025fast}. We train Diffusion Policy from scratch for 100k steps conditioned on a 4-timestep history of proprioception and 448px RGB observations to iteratively denoise 16 future absolute end-effector poses in SE(3). We finetune $\pi_0$-FAST for 30k steps using Low Rank Adaptation (LoRA)~\cite{lora} (rank=16), which takes a single 224px square image (to match pretraining resolution) and predicts a 10-step relative joint angle action-chunk. Training the diffusion policy takes approximately 3 hours on a single NVIDIA GH200, while $\pi_0$-FAST finetuning takes 11 hours. At deployment, both models receive raw RGB images and robot proprioception—SE(3) absolute end-effector pose for diffusion and joint positions for $\pi_0$-FAST—and output the corresponding action targets. To improve temporal consistency between actions predicted at different timesteps, we apply temporal ensembling~\cite{zhao2023learning} to predicted action-chunks during execution for both models. More training details can be found in \cref{sec:appendix_extended_training}.

\figAvgPerf{t!}

\section{Experiments}
We conduct 1,050 physical robot evaluations on an ABB YuMi IRB14000 Bimanual Robot (a robot embodiment unseen during $\pi_0$-FAST pre-training) across five manipulation tasks using the trained policies. Policies are trained on either human teleoperation data or synthetic demonstrations generated by \algabbr. To assess how policy performance scales with training data, we train models with 50, 100, 150, and 1,000 rendered trajectories and up to 150 teleoperation trajectories per task. To ensure a fair comparison, all models are trained for a fixed number of training steps using only third-person RGB observations.

We deliberately selected tasks to highlight R2R2R’s ability to scale across diverse manipulation scenarios that involve varying physical and kinematic structures. Specifically, the tasks span: single-object picking (\textit{``pick up the toy tiger”}), multi-object interaction (\textit{``put the mug on the coffee maker”}), articulated object manipulation (\textit{``turn the faucet off”} and \textit{``open the drawer”}), and bimanual coordination (\textit{``pick up the package with both hands”}). These categories correspond directly to \algabbr’s support for part-level segmentation, articulated object reconstruction, and multi-arm grasp planning, and are visualized in appendix \cref{sec:appendix_vis_coffeee,sec:appendix_vis_faucet,sec:appendix_vis_drawer,sec:appendix_vis_bimanual_lift,sec:appendix_vis_tiger}. We provide additional ablation experiments on trajectory interpolation (\cref{sec:appendix_traj_interp}), increased background randomization (\cref{sec:appendix_background_aug_ablation}), and sim-real co-training (\cref{sec:appendix_sim_real_co_training}) in the appendix.

\subsection{Performance Scaling and Comparison}

To evaluate how well \algabbr-generated data supports policy learning compared to human teleoperated data, we analyze performance trends as a function of dataset size across the five tasks described above. Results are summarized in ~\cref{fig:avg_performance_vs_time}. We observe that \algabbr-generated data scales predictably with dataset size: success rates increase monotonically for most tasks as the number of demonstrations grows. On the “Put the mug on the coffee maker” task (see \cref{fig:avg_performance_vs_time}b), performance of Diffusion Policy trained on \algabbr data improves from 33.3\% at 150 demos to 53.3\% at 1000 demos, while $\pi_0$-FAST jumps from 33.3\% to 80.0\%. While higher quality, real-world data offers better performance in low-data regimes (e.g., $\pi_0$-FAST reaches 73.3\% at 150 real demos vs. 33.3\% at 150 \algabbr demos as shown in~\cref{fig:avg_performance_vs_time}b), as the scale increases to 1000 demos, \algabbr achieves performance that matches or surpasses teleoperation across multiple tasks. This suggests that while real data is more efficient per demonstration, \algabbr's generation enables scaling \textit{trajectory diversity} far beyond human throughput, achieving competitive final performance with less collection effort.

To assess whether this performance is statistically comparable, we conduct formal significance and equivalence testing across all tasks and models. Appendix~\ref{sec:appendix_p_test} shows that on the evaluated tasks, there are no statistically significant differences between policies trained on \algabbr versus human teleoperation data on the tasks we evaluated. Two One-Sided Tests (TOST) further suggest that the observed differences fall within a $\pm$5\% margin, indicating similar overall performance. These findings suggest that \algabbr may offer a viable and scalable alternative to human teleoperation for training effective robot policies.

\section{Conclusion}

We propose \algabbr, a scalable data generation pipeline that creates robot training data from an object scan and a human demonstration video. \algabbr mitigates limitations of prior work by removing the need for teleoperation, robot hardware, or dynamics simulation. It leverages 3D Gaussian Splatting to represent both rigid and articulated objects, enabling parallel rendering using Gaussian-converted meshes and scalable rendering engines. These realistic renderings serve as visual observations for policy training. Given the robot's URDF, \algabbr synthesizes diverse robot trajectories with extracted object motion from one human demonstration using differential inverse kinematics. Experiments on five robotic tasks suggest that policies trained on data generated by \algabbr scale with data volume and perform comparably to those trained on teleoperated demonstrations, demonstrating that \algabbr is a practical and scalable pipeline for real-world robot dexterous manipulation policy learning.

\section{Limitations}
Real2Render2Real (R2R2R) enables scalable data generation and competitive real-world performance, but several limitations remain.

\textbf{Reconstruction and Simulation Fidelity.}
R2R2R relies on vision-based reconstruction methods—such as 3D Gaussian Splatting and mesh conversion—that yield high-fidelity appearance but often lack watertight or physically plausible geometry. These limitations make it difficult to simulate realistic physical interactions, especially in contact-rich settings. As a result, R2R2R forgoes physics simulation entirely. While this design choice boosts scalability, it also restricts modeling of important dynamics such as friction, compliance, and force feedback. As real-to-sim pipelines mature in their physical realism~\cite{pfaff2025_scalable_real2sim}, future versions of R2R2R could integrate simulation layers to better support tasks like slip detection, in-hand correction, or deformable object interaction.

\textbf{Scene Diversity and Collision Awareness.}
Trajectory generation in R2R2R is performed via geometric interpolation, without considering environmental context such as distractor objects or obstacles. As a result, synthesized trajectories may intersect with the scene geometry, leading to physically infeasible plans. Incorporating fast motion planning techniques during trajectory synthesis could improve collision avoidance and robustness, particularly in cluttered or multi-object scenes.

\textbf{Scope of Manipulation Tasks.}
The current framework focuses exclusively on rigid and articulated objects using prehensile manipulation. It does not support deformable object handling or non-prehensile strategies such as pushing, toppling, or sliding. These interactions often demand accurate metric depth estimates and fine-grained physical modeling—both of which are challenging with monocular video and approximate geometry. Extending R2R2R to these broader manipulation regimes remains an open direction.

\textbf{Grasping Generality.}
R2R2R's grasp generation module currently uses antipodal grasp sampling, which limits compatibility to parallel-jaw grippers. This restricts the generality of trained policies and excludes multi-fingered or anthropomorphic hands, which require richer grasp representations and contact models. Supporting these more complex end-effectors would require advances in grasp synthesis and simulation.

\textbf{Tracking Robustness.}
Like other object-centric pipelines, R2R2R is vulnerable to tracking failures under fast motion, heavy occlusion, poor texture, or reflective surfaces. In such cases, object reconstructions and pose tracks may be inaccurate, resulting in degraded data quality. These failures can lead to invalid grasps or trajectories that do not transfer well to the real world. Robustifying tracking and adding confidence-aware filtering or correction is an important area for future work.

Addressing these limitations—through richer physical modeling, context-aware planning, expanded manipulation capabilities, and improved reconstruction robustness—offers a plausible path toward more general and reliable robot learning at scale.

\section{Acknowledgement}
This research was performed at the AUTOLAB at UC Berkeley in affiliation with the Berkeley AI Research (BAIR) Lab, and the CITRIS "People and Robots" (CPAR) Initiative.
In their academic roles at UC Berkeley, Justin Yu, Letian Fu, Huang Huang, Karim El-Refai, and Ken Goldberg are supported in part by donations from Toyota Research Institute, Autodesk, Meta, Google, Siemens, Bosch, and by equipment grants from NVIDIA, PhotoNeo, NSF AI4OPT Centre, and Intuitive Surgical. This material is based upon work supported by the National Science Foundation Graduate Research Fellowship Program under Grant No. DGE 2146752. Any opinions, findings, and conclusions or recommendations expressed in this material are those of the author(s) and do not necessarily reflect the views of the National Science Foundation. We deeply appreciate Jon Kuroda, Ion Stoica, and Joey Gonzalez for their generous support with computing hardware.

\newpage
\section{Appendix}
\etocsettocstyle{\section*{Contents for This Section}}{}
\localtableofcontents

\newpage
\subsection{Raw Evaluation Results}
\label{sec:eval_results}
We report raw task success rates for each policy in \cref{tab:combined_eval}.

\begin{table*}[h!]
    \small
    \centering
    \begin{tabular}{lccc|cccc}
        \toprule
        \multirow{2}{*}{\textbf{Task / Policy}} & \multicolumn{3}{c|}{\textbf{Teleop Trajectories}} & \multicolumn{4}{c}{\textbf{R2R2R Trajectories}} \\
         & 50 & 100 & 150 & 50 & 100 & 150 & 1000 \\
        \midrule
        \textbf{Pick up the tiger} & & & & & & & \\
        Diffusion Policy & 6.7\% & 66.7\% & 73.3\% & 0.0\% & 26.7\% & 40.0\% & 66.6\% \\
        $\pi_0$-FAST (Finetuned) & 26.7\% & 40.0\% & 73.3\% & 0.0\% & 0.0\% & 6.7\% & 80.0\% \\
        \midrule
        
        \textbf{Put the mug on the coffee maker} & & & & & & & \\
        Diffusion Policy & 13.3\% & 33.3\% & 40.0\% & 13.3\% & 13.3\% & 33.3\% & 53.3\% \\
        $\pi_0$-FAST (Finetuned) & 6.6\% & 13.3\% & 73.3\% & 0.0\% & 0.0\% & 33.3\% & 80.0\% \\
        \midrule
        
        \textbf{Pick up the package with both hands} & & & & & & & \\
        Diffusion Policy & 66.7\% & 66.7\% & 80.0\% & 20.0\% & 33.3\% & 20.0\% & 73.3\% \\
        $\pi_0$-FAST (Finetuned) & 6.7\% & 46.7\% & 60.0\% & 6.6\% & 13.3\% & 6.6\% & 66.7\% \\
        \midrule
        
        \textbf{Open the drawer} & & & & & & & \\
        Diffusion Policy & 20.0\% & 60.0\% & 66.7\% & 13.3\% & 33.3\% & 46.7\% & 66.7\% \\
        $\pi_0$-FAST (Finetuned) & 0.0\% & 40.0\% & 60.0\% & 0.0\% & 20.0\% & 13.3\% & 86.6\% \\
        \midrule
        
        \textbf{Turn the faucet off} & & & & & & & \\
        Diffusion Policy & 20.0\% & 46.7\% & 66.7\% & 20.0\% & 33.3\% & 53.3\% & 80.0\% \\
        $\pi_0$-FAST (Finetuned) & 35.3\% & 60.0\% & 80.0\% & 0.0\% & 13.3\% & 13.3\% & 80.0\% \\
        \bottomrule
    \end{tabular}
    \caption{\textbf{Comparison of Physical Policy Success Rates Across Training Sources.} Task success rates for Diffusion Policy and $\pi_0$-FAST trained exclusively on either human teleoperation data (left) or R2R2R-generated data (right). Each policy was evaluated on 15 trials per task using a binary success metric: a score of 1 is assigned for successful task completion, and 0 otherwise.}
    \label{tab:combined_eval}
\end{table*}
\newpage

\subsection{Additional Ablation Experiments}
\subsubsection{Trajectory Interpolation}
\label{sec:appendix_traj_interp}

R2R2R generates diverse trajectories by adapting a single human demonstration to new object poses through interpolation and spatial transformation (see \cref{ssec:real_sim_traj_extrac} and \cref{fig:traj_interp} for visualization). To evaluate the impact of this trajectory interpolation step, we ablate it by replaying only the original object motion track without adapting to varied initial and goal poses. Table~\ref{tab:traj_interp_ablation} shows a substantial drop in performance when interpolation is disabled: on the ``Put the mug on the coffee maker'' task, success rates fall from 80.0\% to 0.0\% for $\pi_0$-FAST and from 53.3\% to 6.7\% for Diffusion Policy. This highlights that simply replaying object motion from a single demonstration is insufficient for generating transferable robot behaviors—trajectory adaptation is crucial to scaling data diversity in object-centric manipulation.

\begin{table}[h]
\centering
\begin{tabular}{lcc}
\toprule
Policy & w/o Trajectory Interpolation (1k) & w/ Trajectory Interpolation (1k) \\
\midrule
$\pi_0$-FAST (Finetuned) & 0.0\% & \textbf{80.0\%} \\
Diffusion Policy & 6.7\% & \textbf{53.3\%} \\
\bottomrule
\end{tabular}
\vspace{4pt}
\caption{Success rates on ``Put the mug on the coffee maker'' using R2R2R-generated data with and without trajectory interpolation (1,000 demos). Interpolation enables adapting object motion to varied contexts, which is critical for policy generalization.}
\label{tab:traj_interp_ablation}
\vspace{-1.5em}
\end{table}

\subsubsection{Background and Tabletop Texture Augmentation}
\label{sec:appendix_background_aug_ablation}
\begin{figure}[h!]
    \centering
    \begin{tabular}{ccc}
        \includegraphics[width=0.30\textwidth]{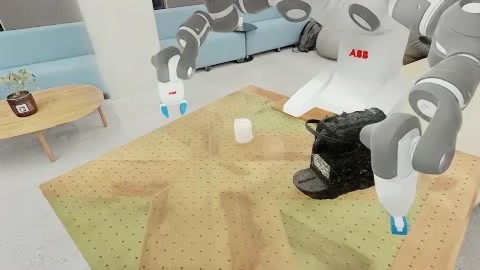} &
        \rotatebox[origin=c]{180}{\includegraphics[width=0.30\textwidth]{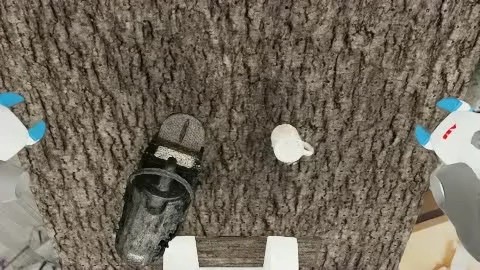}} &
        \includegraphics[width=0.30\textwidth]{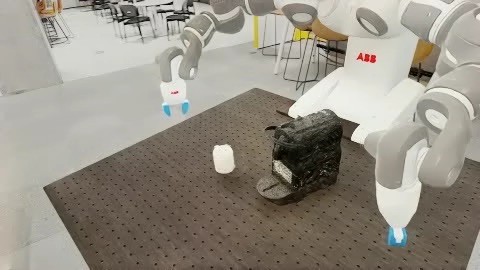} \\
        \rotatebox[origin=c]{180}{\includegraphics[width=0.30\textwidth]{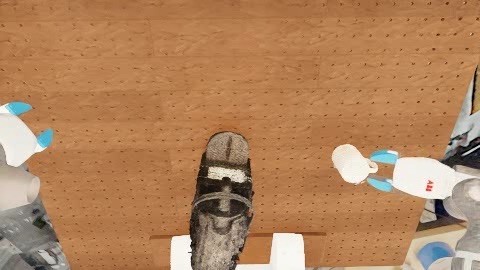}} &
        \includegraphics[width=0.30\textwidth]{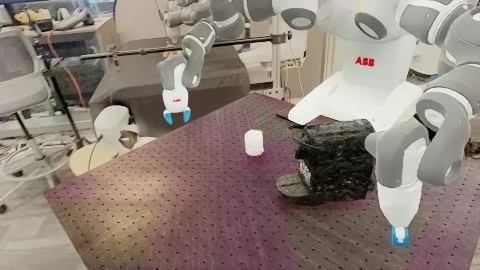} &
        \rotatebox[origin=c]{180}{\includegraphics[width=0.30\textwidth]{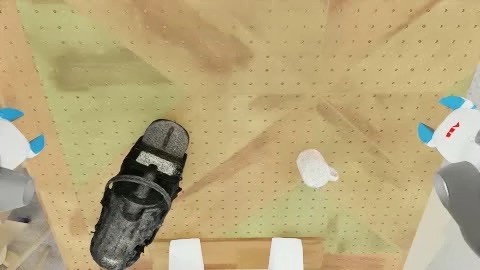}} \\
    \end{tabular}
    \caption{\textbf{Background and Tabletop Texture Augmentation} -- Each image corresponds to a different environment.}
    \label{fig:background_abl}
\end{figure}

Our default data generation pipeline includes moderate visual augmentation, such as randomized lighting, camera pose, and object placement, as well as sampling from a limited set of lightbox-style background environments. To study the effect of stronger visual perturbations, we apply more aggressive augmentation that includes a wider variety of lightbox backgrounds and diverse tabletop textures (see \cref{fig:background_abl}).

\Cref{tab:background} reports the success rates on the task \emph{Put the mug on the coffee maker} under this more varied visual setting. We observe a consistent drop in policy performance across both $\pi_0$-FAST and Diffusion Policy when trained on data with aggressive background and surface augmentation. This result suggests that while visual diversity is generally beneficial, overly strong appearance perturbations may harm policy learning when not properly balanced. Future work may investigate more principled augmentation schedules or adaptive augmentation strategies to preserve generalization while maintaining performance.

\begin{table}[h]
\centering
\begin{tabular}{lcc}
\toprule
Policy & Less Background Aug. (1k) & More Background Aug. (1k) \\
\midrule
$\pi_0$-FAST (Fine-tuned) & \textbf{73.3}\% & 35.3\% \\
Diffusion Policy & \textbf{53.3\%} & 33.3\% \\
\bottomrule
\end{tabular}
\vspace{5.0pt}
\caption{Success rate comparison on \emph{Put the mug on the coffee maker} with and without background and tabletop texture augmentation. We generated 1000 trajectories for each setting and evaluated across 15 trials.}
\label{tab:background}
\end{table}

\newpage
\subsubsection{Sim-and-Real Co-training}
\label{sec:appendix_sim_real_co_training}
While sim-and-real co-training is not the main focus of this paper, we included additional results comparing policies exclusively on either R2R2R-generated data, human teleoperation data, and co-training setup that combines data from both sources. Specifically, for the task \emph{Put the mug on the coffee maker}, we trained a policy using 1,000 R2R2R-generated demonstrations together with 150 human teleoperation demonstrations. We do not perform additional importance sampling or re-weighting of human teleoperation data. For the $\pi_0$-FAST policy, co-training achieved a success rate of 73.3\%, which is on par with training using only R2R2R data or only real demonstrations individually. Co-training for diffusion policy yields a significant improvement over either real data only or R2R2R-generated data only, where the performance improved from 40.0\% to 86.7\%. We hypothesize that since LoRA~\cite{lora} serves as a significant regularizer for $\pi_0$-FAST, end-to-end fine-tuning with completely unfrozen model with additional hyperparameters tuning could lead to better performance. For more in-depth analysis on how co-training can improve policy performance, please refer to~\cite{maddukuri2025simandreal, wei2025empirical}.

\begin{table}[h]
\centering
\begin{tabular}{lccc}
\toprule
Policy & Real Data Only (150) & R2R2R Data Only (1k) & Co-Training (150+1k) \\
\midrule
$\pi_0$-FAST & 73.3\% & \textbf{80.0\%} & 73.3\% \\
Diffusion Policy & 40.0\% & 53.3\% & \textbf{86.7\%} \\
\bottomrule
\end{tabular}
\vspace{5.0pt}
\caption{Success rate comparison on \emph{Put the mug on the coffee maker} under different training datasets mixtures.}
\label{tab:co_training}
\end{table}

\newpage
\subsection{Task Visualizations}
Physical policy rollout figures show model input RGB frames from real policy evaluation successes using either Diffusion Policy~\cite{chi2023diffusion} or $\pi_0$-FAST~\cite{pertsch2025fast}. The depicted policies were trained \textit{exclusively} on R2R2R synthetic data.
\subsubsection{Put the Mug on the Coffee Maker}
\label{sec:appendix_vis_coffeee}

\begin{figure}[h!]
    \centering
    \begin{tabular}{ccc}
        \includegraphics[width=0.30\textwidth]{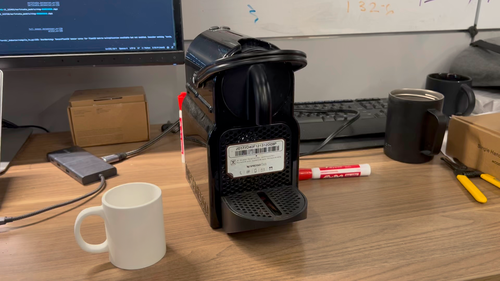} &
        \includegraphics[width=0.30\textwidth]{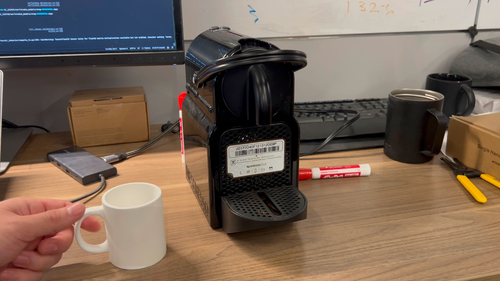} &
        \includegraphics[width=0.30\textwidth]{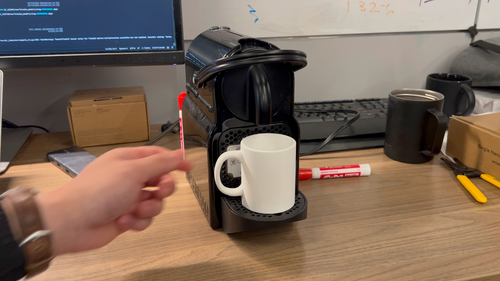} 
    \end{tabular}
    \caption{\textbf{Put the Mug on the Coffee Maker -- Demonstration Video Frames}.}
    \label{fig:faucet_demo}
\end{figure}

\begin{figure}[h!]
    \centering
    \begin{tabular}{ccc}
        \rotatebox[origin=c]{180}{\includegraphics[width=0.30\textwidth]{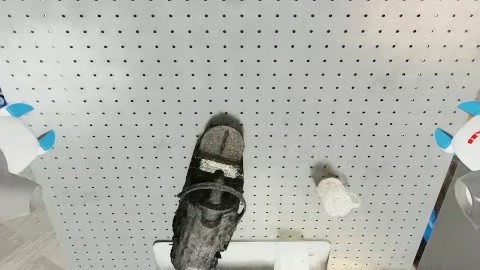}} &
        \rotatebox[origin=c]{180}{\includegraphics[width=0.30\textwidth]{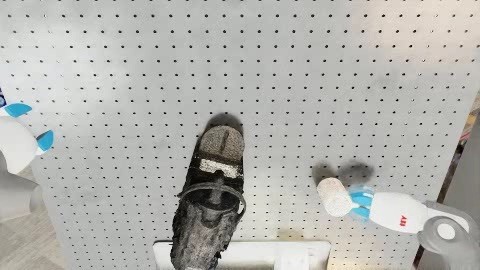}} &
        \rotatebox[origin=c]{180}{\includegraphics[width=0.30\textwidth]{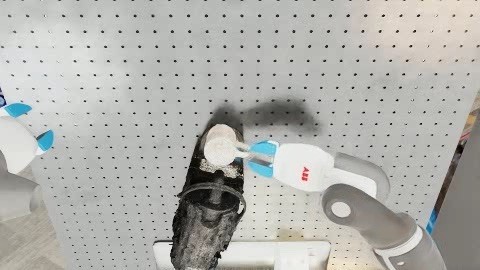}} \\
        \includegraphics[width=0.30\textwidth]{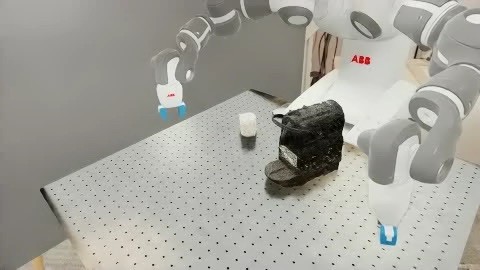} &
        \includegraphics[width=0.30\textwidth]{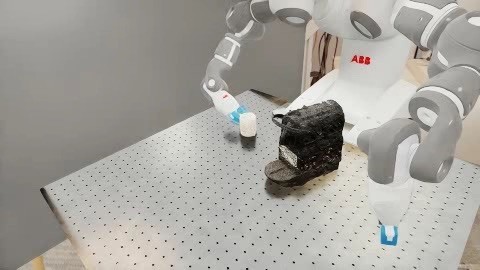} &
        \includegraphics[width=0.30\textwidth]{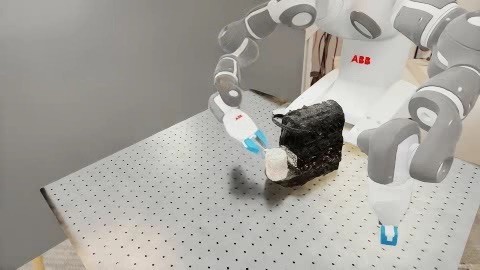} \\
    \end{tabular}
    \caption{\textbf{Put the Mug on the Coffee Maker -- Example R2R2R Frames}.}
    \label{fig:coffee_maker_r2r2r}
\end{figure}

\begin{figure}[h!]
    \centering
    \begin{tabular}{ccc}
        \rotatebox[origin=c]{180}{\includegraphics[width=0.30\textwidth]{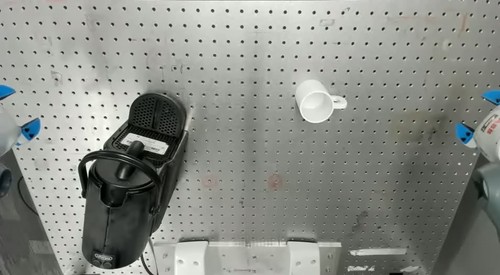}} &
        \rotatebox[origin=c]{180}{\includegraphics[width=0.30\textwidth]{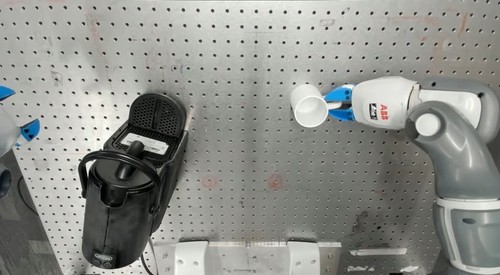}} &
        \rotatebox[origin=c]{180}{\includegraphics[width=0.30\textwidth]{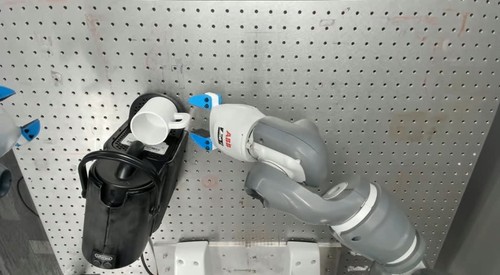}} \\
        \includegraphics[width=0.30\textwidth]{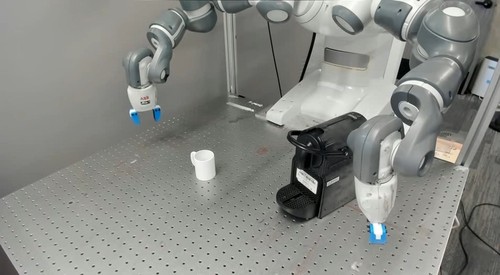} &
        \includegraphics[width=0.30\textwidth]{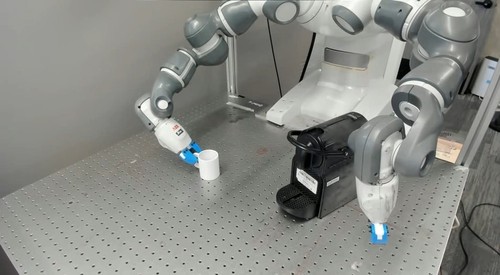} &
        \includegraphics[width=0.30\textwidth]{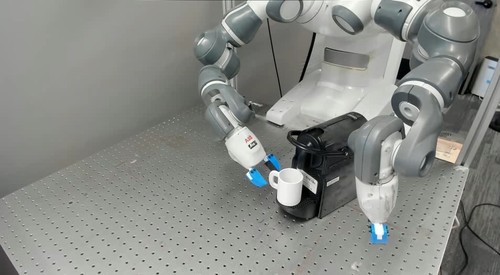} \\
    \end{tabular}
    \caption{\textbf{Put the Mug on the Coffee Maker -- Physical Policy Rollout}.}
    \label{fig:coffee_maker_real}
\end{figure}

\newpage
\subsubsection{Turn off the Faucet}
\label{sec:appendix_vis_faucet}

\begin{figure}[h!]
    \centering
    \begin{tabular}{ccc}
        \includegraphics[width=0.30\textwidth]{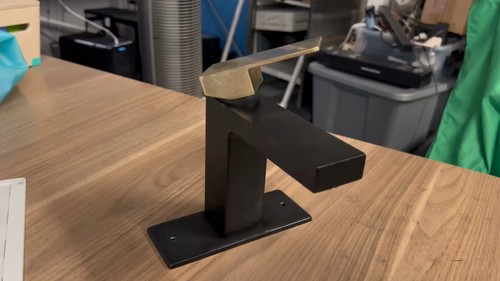} &
        \includegraphics[width=0.30\textwidth]{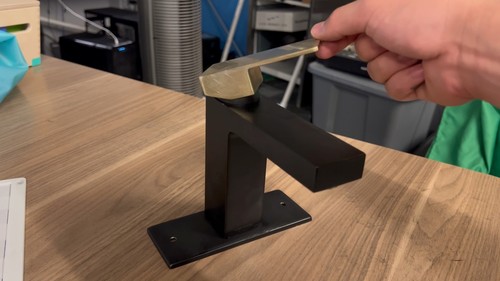} &
        \includegraphics[width=0.30\textwidth]{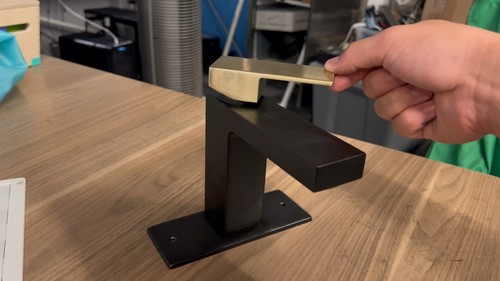} 
    \end{tabular}
    \caption{\textbf{Turn off the Faucet - Demonstration Video Frames}.}
    \label{fig:faucet_demo}
\end{figure}

\begin{figure}[h!]
    \centering
    \begin{tabular}{ccc}
        \rotatebox[origin=c]{180}{\includegraphics[width=0.30\textwidth]{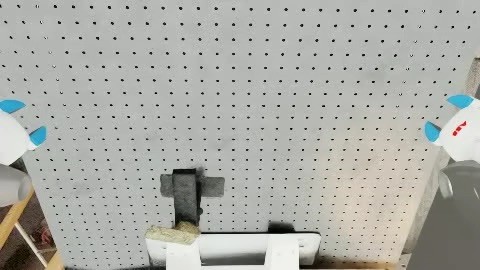}} &
        \rotatebox[origin=c]{180}{\includegraphics[width=0.30\textwidth]{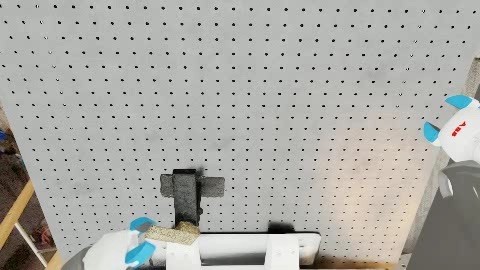}} &
        \rotatebox[origin=c]{180}{\includegraphics[width=0.30\textwidth]{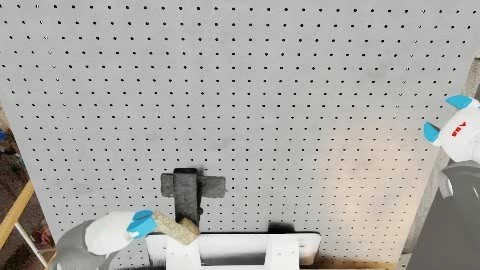}} \\
        \includegraphics[width=0.30\textwidth]{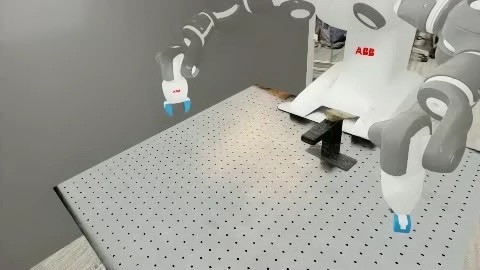} &
        \includegraphics[width=0.30\textwidth]{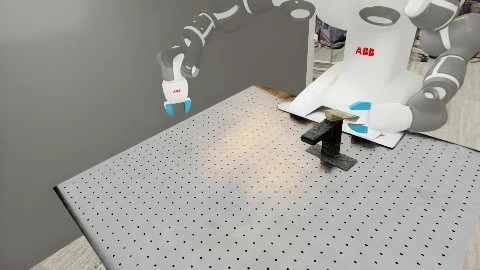} &
        \includegraphics[width=0.30\textwidth]{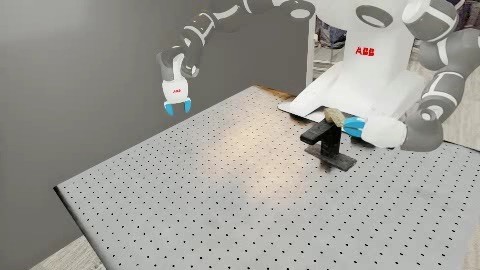} \\
    \end{tabular}
    \caption{\textbf{Turn off the Faucet - Example R2R2R Frames}.}
    \label{fig:faucet_r2r2r}
\end{figure}

\begin{figure}[h!]
    \centering
    \begin{tabular}{ccc}
        \rotatebox[origin=c]{180}{\includegraphics[width=0.30\textwidth]{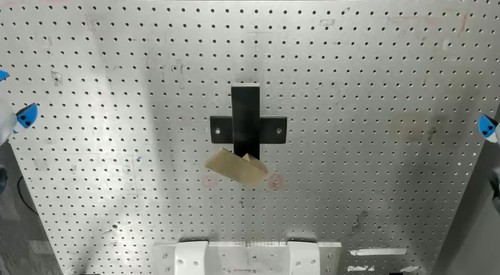}} &
        \rotatebox[origin=c]{180}{\includegraphics[width=0.30\textwidth]{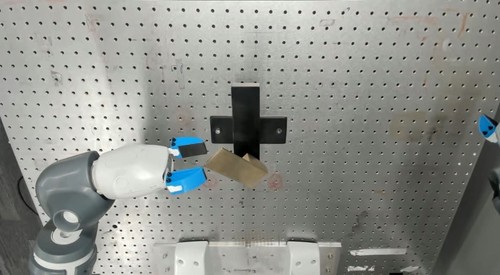}} &
        \rotatebox[origin=c]{180}{\includegraphics[width=0.30\textwidth]{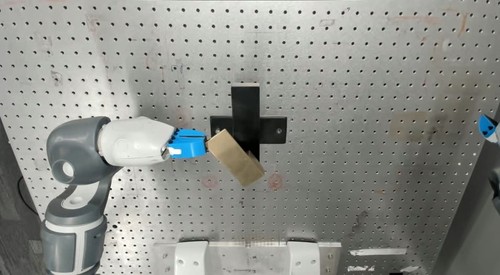}} \\
        \includegraphics[width=0.30\textwidth]{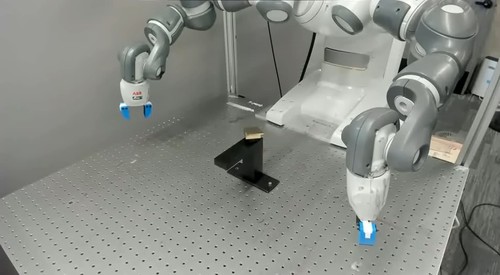} &
        \includegraphics[width=0.30\textwidth]{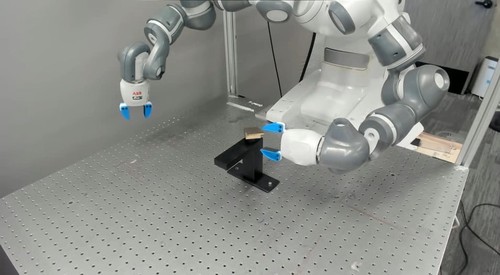} &
        \includegraphics[width=0.30\textwidth]{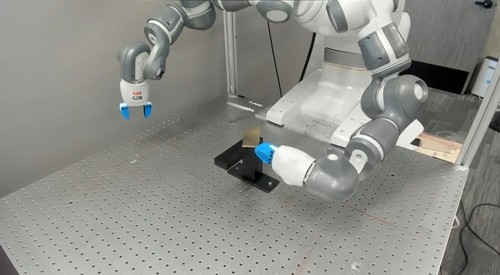} \\
    \end{tabular}
    \caption{\textbf{Turn off the Faucet - Physical Policy Rollout}.}
    \label{fig:faucet_real}
\end{figure}

\textbf{Note:} For human teleoperated demonstrations, the teleoperator would push down on the faucet handle in a non-prehensile motion to turn it off instead of grasping the handle and twisting it closed as is done with \algabbr--where only prehensile grasping is currently supported. 

\newpage

\subsubsection{Open the Drawer}
\label{sec:appendix_vis_drawer}

\begin{figure}[h!]
    \centering
    \begin{tabular}{ccc}
        \includegraphics[width=0.30\textwidth]{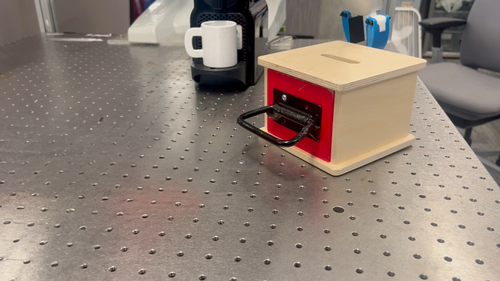} &
        \includegraphics[width=0.30\textwidth]{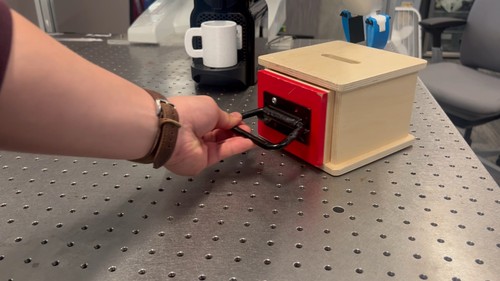} &
        \includegraphics[width=0.30\textwidth]{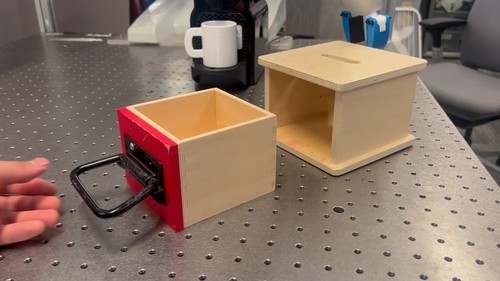} 
    \end{tabular}
    \caption{\textbf{Open the Drawer - Demonstration Video Frames}. \textbf{Note:} The true video order--and thus the tracked trajectory--was in reverse, as a full multi-view scan for the inner drawer requires it to first be in an open configuration.}
    \label{fig:drawer_demo}
\end{figure}

\begin{figure}[h!]
    \centering
    \begin{tabular}{ccc}
        \rotatebox[origin=c]{180}{\includegraphics[width=0.30\textwidth]{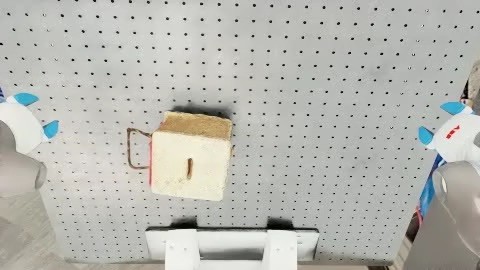}} &
        \rotatebox[origin=c]{180}{\includegraphics[width=0.30\textwidth]{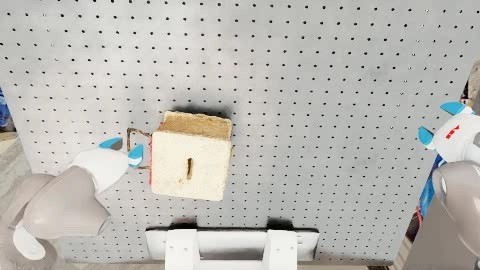}} &
        \rotatebox[origin=c]{180}{\includegraphics[width=0.30\textwidth]{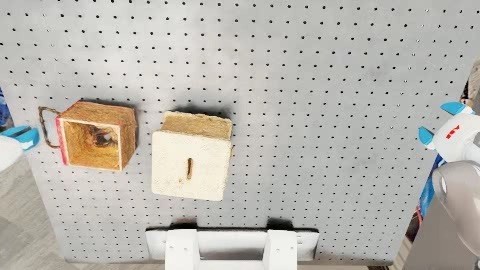}} \\
        \includegraphics[width=0.30\textwidth]{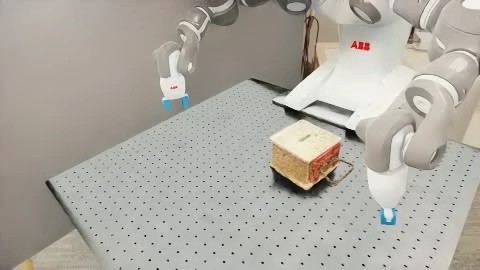} &
        \includegraphics[width=0.30\textwidth]{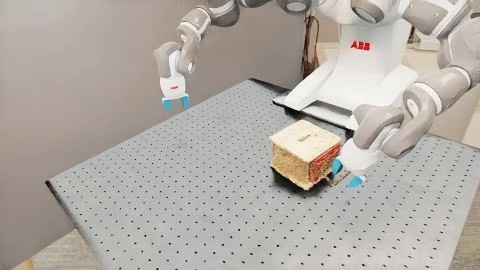} &
        \includegraphics[width=0.30\textwidth]{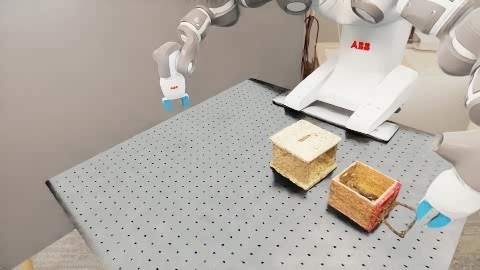} \\
    \end{tabular}
    \caption{\textbf{Open the Drawer - Example R2R2R Frames}.}
    \label{fig:drawer_r2r2r}
\end{figure}

\begin{figure}[h!]
    \centering
    \begin{tabular}{ccc}
        \rotatebox[origin=c]{180}{\includegraphics[width=0.30\textwidth]{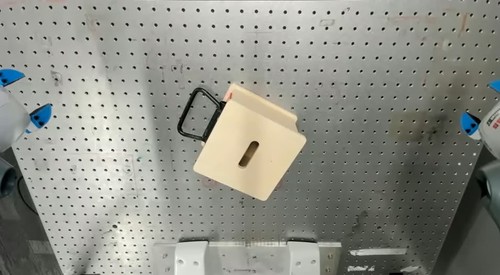}} &
        \rotatebox[origin=c]{180}{\includegraphics[width=0.30\textwidth]{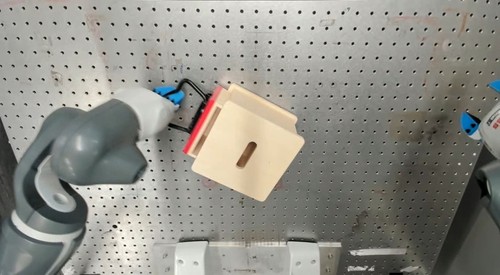}} &
        \rotatebox[origin=c]{180}{\includegraphics[width=0.30\textwidth]{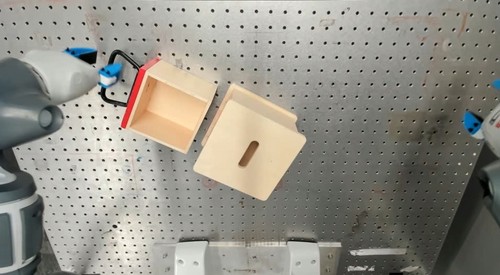}} \\
        \includegraphics[width=0.30\textwidth]{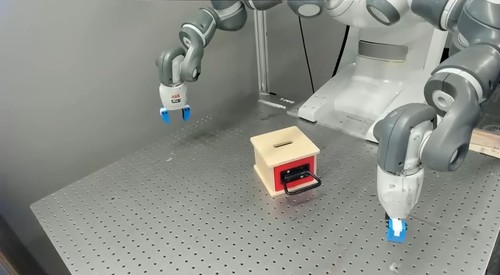} &
        \includegraphics[width=0.30\textwidth]{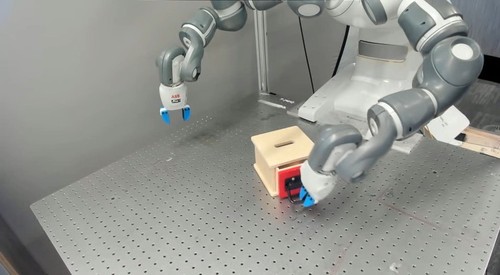} &
        \includegraphics[width=0.30\textwidth]{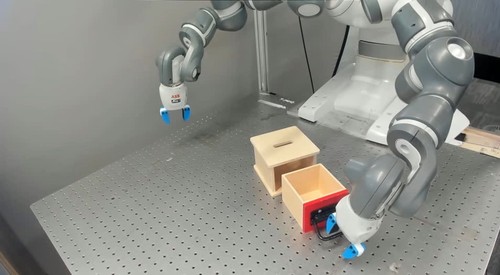} \\
    \end{tabular}
    \caption{\textbf{Open the Drawer - Physical Policy Rollout}.}
    \label{fig:drawer_real}
\end{figure}

\newpage

\subsubsection{Lift up the Package with Both Hands}
\label{sec:appendix_vis_bimanual_lift}

\begin{figure}[h!]
    \centering
    \begin{tabular}{ccc}
        \includegraphics[width=0.30\textwidth]{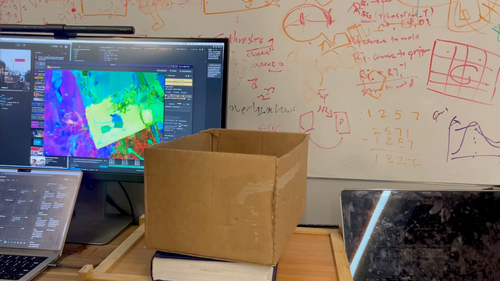} &
        \includegraphics[width=0.30\textwidth]{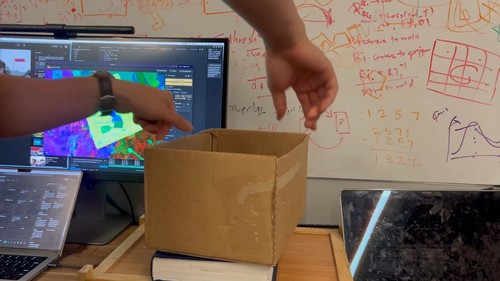} &
        \includegraphics[width=0.30\textwidth]{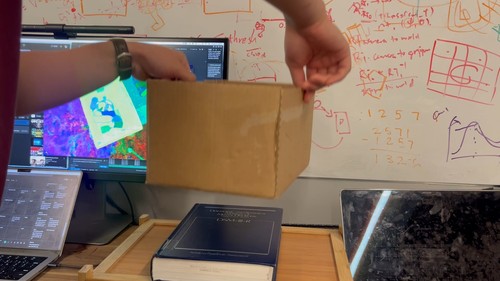} 
    \end{tabular}
    \caption{\textbf{Lift up the Package with Both Hands - Demonstration Video Frames}.}
    \label{fig:drawer_demo}
\end{figure}

\begin{figure}[h!]
    \centering
    \begin{tabular}{ccc}
        \rotatebox[origin=c]{180}{\includegraphics[width=0.30\textwidth]{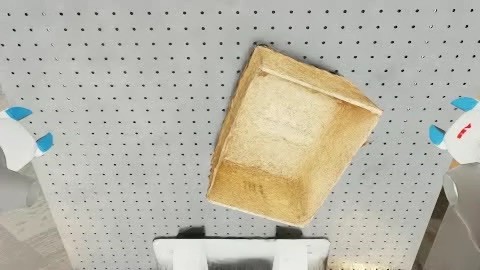}} &
        \rotatebox[origin=c]{180}{\includegraphics[width=0.30\textwidth]{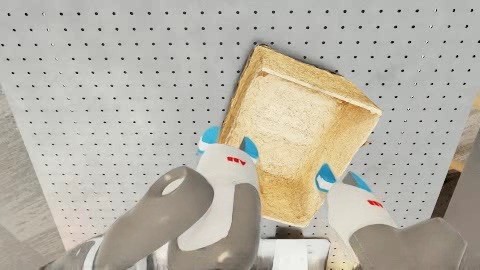}} &
        \rotatebox[origin=c]{180}{\includegraphics[width=0.30\textwidth]{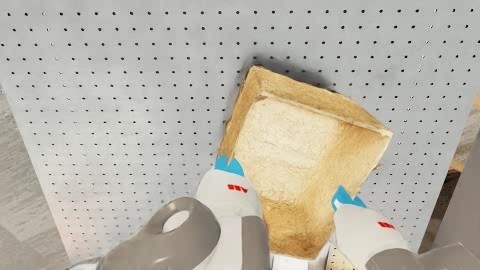}} \\
        \includegraphics[width=0.30\textwidth]{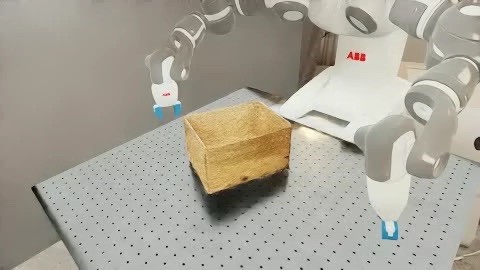} &
        \includegraphics[width=0.30\textwidth]{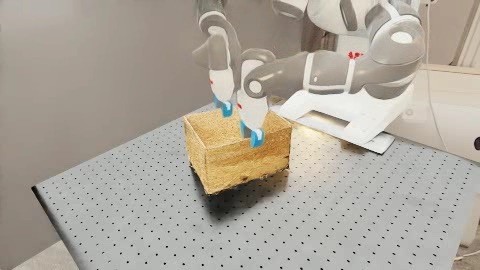} &
        \includegraphics[width=0.30\textwidth]{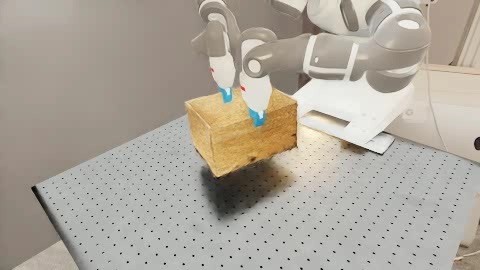} \\
    \end{tabular}
    \caption{\textbf{Lift up the Package with Both Hands - Example R2R2R Frames}.}
    \label{fig:package_r2r2r}
\end{figure}

\begin{figure}[h!]
    \centering
    \begin{tabular}{ccc}
        \rotatebox[origin=c]{180}{\includegraphics[width=0.30\textwidth]{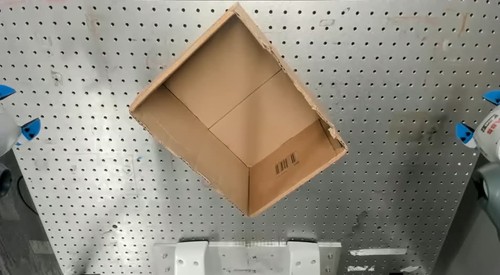}} &
        \rotatebox[origin=c]{180}{\includegraphics[width=0.30\textwidth]{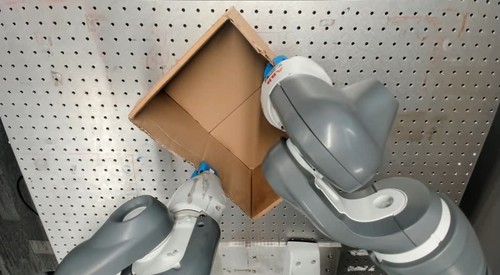}} &
        \rotatebox[origin=c]{180}{\includegraphics[width=0.30\textwidth]{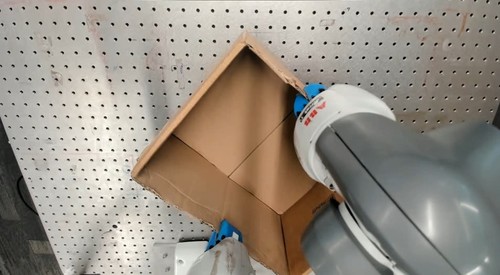}} \\
        \includegraphics[width=0.30\textwidth]{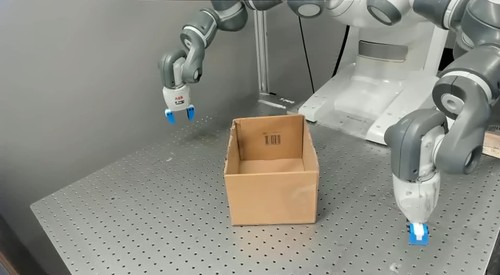} &
        \includegraphics[width=0.30\textwidth]{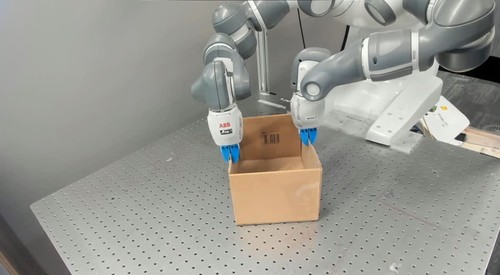} &
        \includegraphics[width=0.30\textwidth]{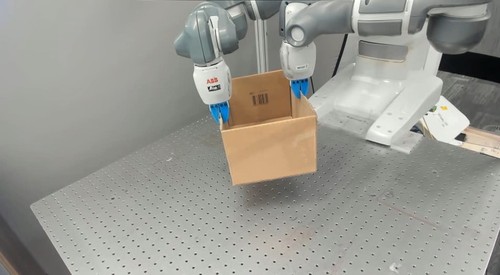} \\
    \end{tabular}
    \caption{\textbf{Lift up the Package with Both Hands - Physical Policy Rollout}.}
    \label{fig:package_real}
\end{figure}

\newpage

\subsubsection{Pick up the Tiger}
\label{sec:appendix_vis_tiger}

\begin{figure}[h!]
    \centering
    \begin{tabular}{ccc}
        \includegraphics[width=0.30\textwidth]{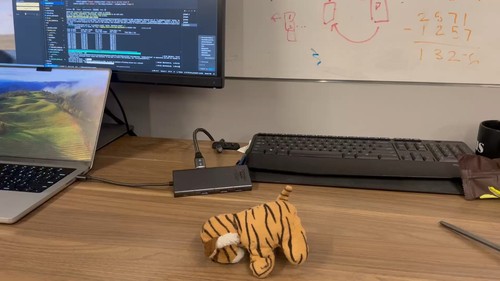} &
        \includegraphics[width=0.30\textwidth]{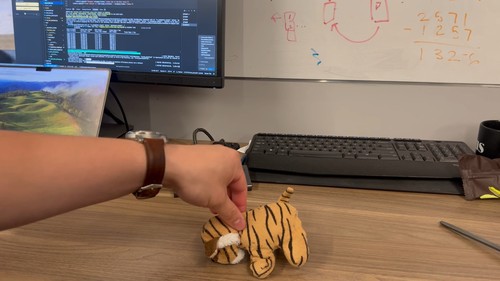} &
        \includegraphics[width=0.30\textwidth]{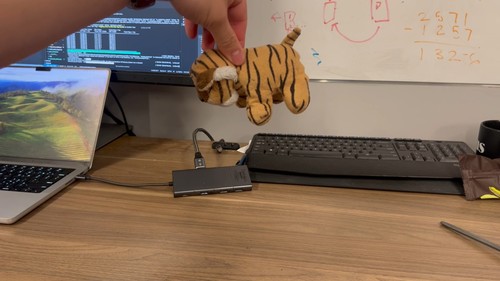} 
    \end{tabular}
    \caption{\textbf{Pick up the Tiger - Demonstration Video Frames}.}
    \label{fig:drawer_demo}
\end{figure}

\begin{figure}[h!]
    \centering
    \begin{tabular}{ccc}
        \rotatebox[origin=c]{180}{\includegraphics[width=0.30\textwidth]{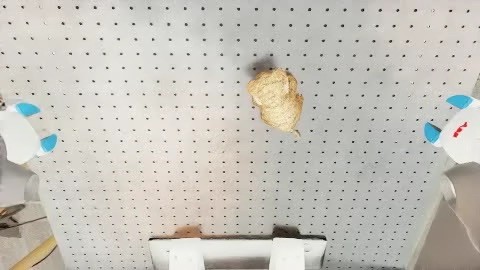}} &
        \rotatebox[origin=c]{180}{\includegraphics[width=0.30\textwidth]{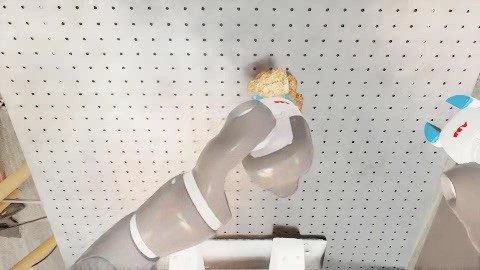}} &
        \rotatebox[origin=c]{180}{\includegraphics[width=0.30\textwidth]{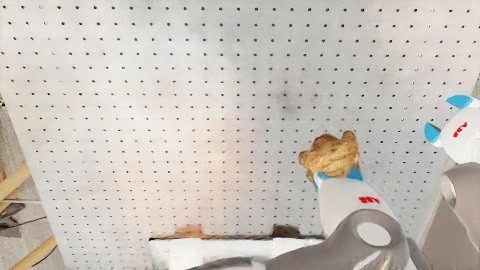}} \\
        \includegraphics[width=0.30\textwidth]{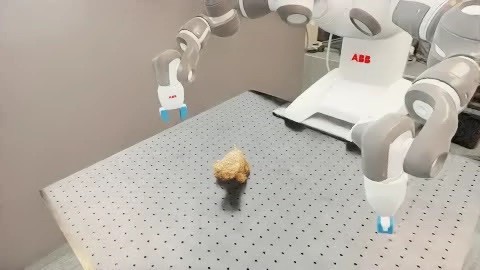} &
        \includegraphics[width=0.30\textwidth]{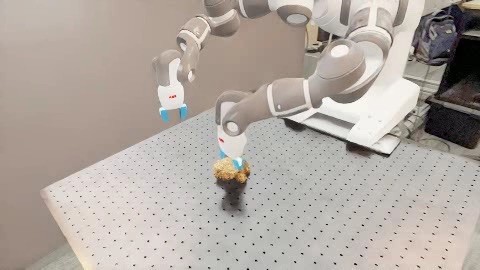} &
        \includegraphics[width=0.30\textwidth]{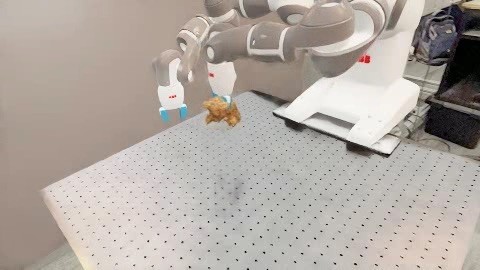} \\
    \end{tabular}
    \caption{\textbf{Pick up the Tiger - Example R2R2R Frames}.}
    \label{fig:tiger_r2r2r}
\end{figure}

\begin{figure}[h!]
    \centering
    \begin{tabular}{ccc}
        \rotatebox[origin=c]{180}{\includegraphics[width=0.30\textwidth]{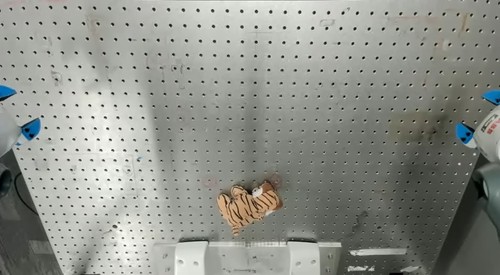}} &
        \rotatebox[origin=c]{180}{\includegraphics[width=0.30\textwidth]{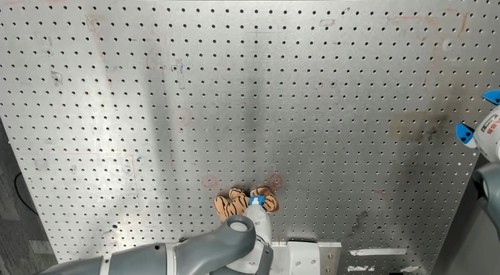}} &
        \rotatebox[origin=c]{180}{\includegraphics[width=0.30\textwidth]{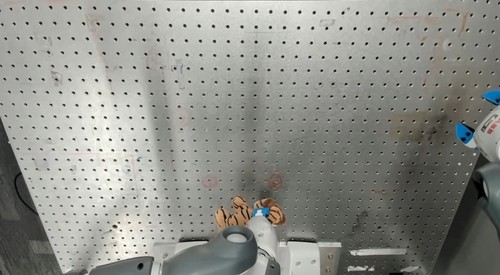}} \\
        \includegraphics[width=0.30\textwidth]{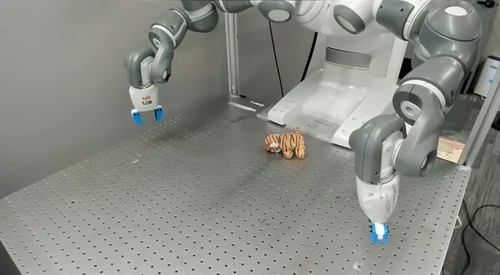} &
        \includegraphics[width=0.30\textwidth]{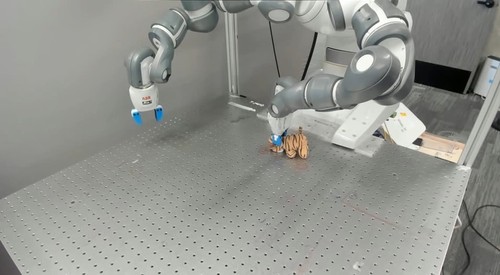} &
        \includegraphics[width=0.30\textwidth]{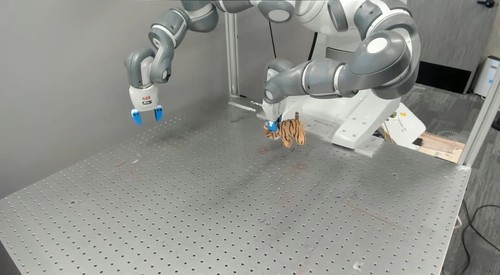} \\
    \end{tabular}
    \caption{\textbf{Pick up the Tiger - Physical Policy Rollout}.}
    \label{fig:tiger_real}
\end{figure}

\newpage

\subsubsection{Put the Mug on the Coffee Maker (Franka Robot Embodiment)}
Unlike vanilla $\pi_0$-FAST (DROID)~\cite{pertsch2025fast}, which operates under joint velocity control, \algabbr records only joint positions. To accommodate this difference, we fine-tuned $\pi_0$-FAST (DROID) to predict delta joint positions (and absolute gripper position, consistent with~\cite{pertsch2025fast}). Since Franka’s impedance control mode can be imprecise, we apply blocking control with temporal ensembling to improve execution accuracy.

Although the learned policy occasionally completes the task successfully, we observe several consistent failure modes, largely stemming from limitations of the default Franka gripper:
\begin{enumerate}
    \item Collision during grasping: The grasp approach that is near parallel to the table often causes the gripper to collide with the table surface during mug pickup.
    \item Off-center grasp: The gripper’s wide jaws tend to produce asymmetric contacts, with one pad closer to the mug than the other. This imbalance induces rotation, leading to slippage.
    \item Difficulty in precise placement: The wide gripper also makes it challenging to release the mug accurately onto the coffee machine.
\end{enumerate}

To mitigate these issues, we recommend using the Robotiq 2F-85 gripper in future experiments. Its smaller form factor and improved grasping precision may reduce failure rates and improve placement consistency.

\begin{figure}[h!]
    \centering
    \begin{tabular}{ccc}
        \includegraphics[width=0.30\textwidth]{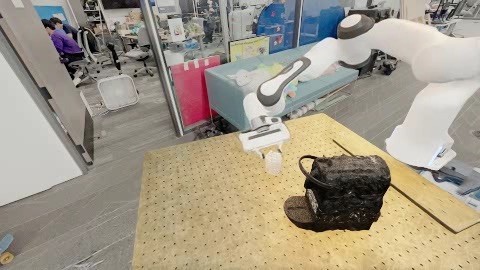} &
        \includegraphics[width=0.30\textwidth]{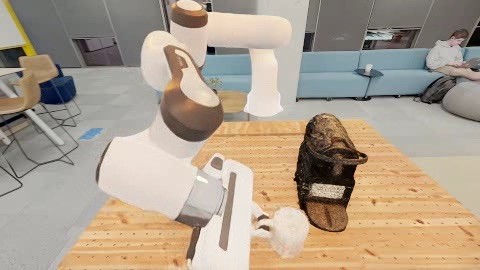} &
        \includegraphics[width=0.30\textwidth]{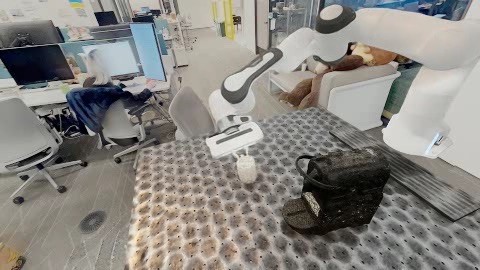} \\
        \includegraphics[width=0.30\textwidth]{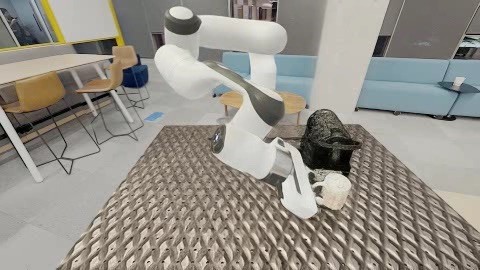} &
        \includegraphics[width=0.30\textwidth]{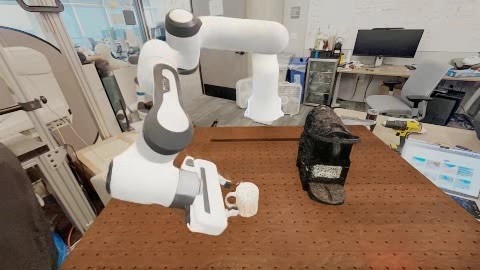} &
        \includegraphics[width=0.30\textwidth]{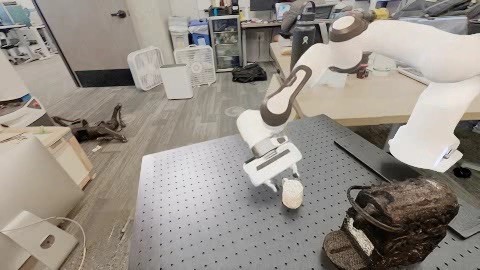} \\
    \end{tabular}
    \caption{\textbf{Put the Mug on the Coffee Maker (Franka Robot) - Example R2R2R Frames}.}
    \label{fig:franka_r2r2r}
\end{figure}

\begin{figure}[h!]
    \centering
    \begin{tabular}{ccc}
        \includegraphics[width=0.30\textwidth]{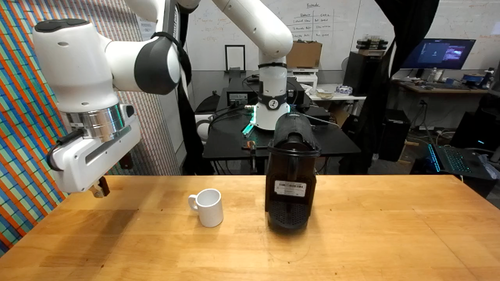} &
        \includegraphics[width=0.30\textwidth]{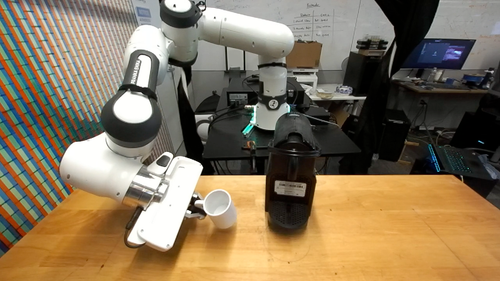} &
        \includegraphics[width=0.30\textwidth]{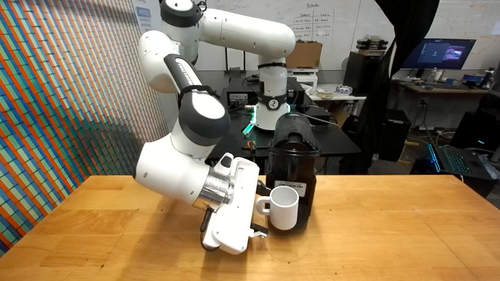} \\
        \includegraphics[width=0.30\textwidth]{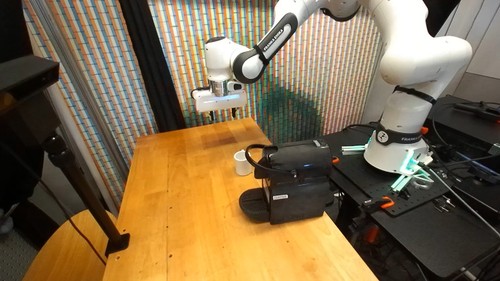} &
        \includegraphics[width=0.30\textwidth]{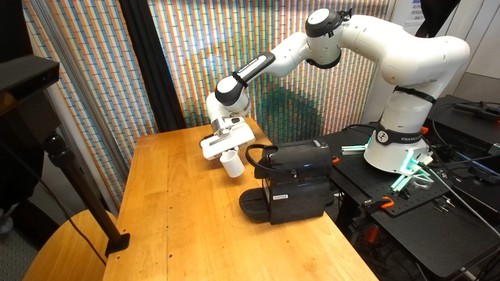} &
        \includegraphics[width=0.30\textwidth]{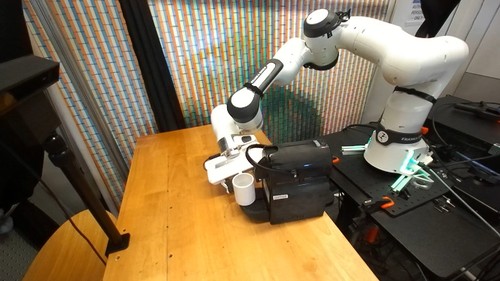} \\
    \end{tabular}
    \caption{\textbf{Put the Mug on the Coffee Maker (Franka Robot) - Physical Policy Rollout}.}
    \label{fig:franka_real}
\end{figure}
\newpage

\subsection{Qualitative Ablations}

\begin{figure}[h!]
    \centering
    \begin{tabular}{ccc}
        \rotatebox[origin=c]{180}{\includegraphics[width=0.30\textwidth]{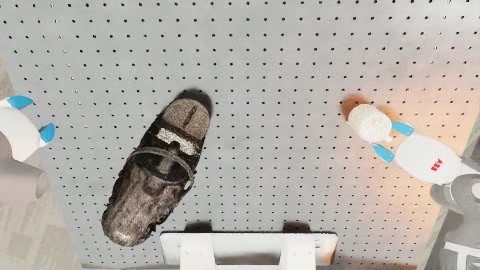}} &
        \rotatebox[origin=c]{180}{\includegraphics[width=0.30\textwidth]{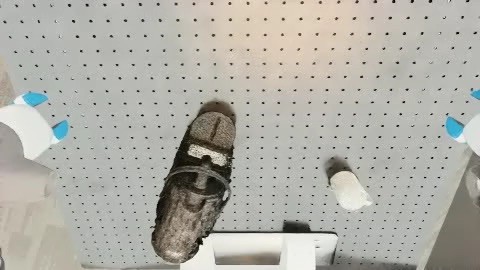}} &
        \rotatebox[origin=c]{180}{\includegraphics[width=0.30\textwidth]{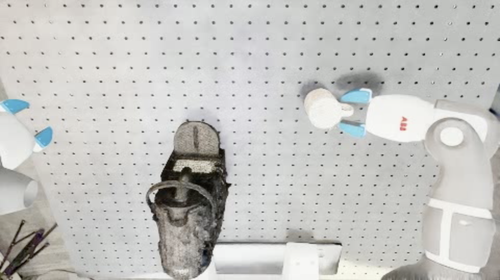}} \\
        \includegraphics[width=0.30\textwidth]{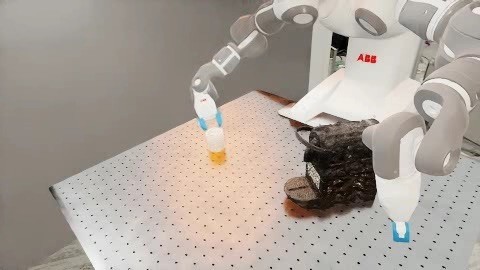} &
        \includegraphics[width=0.30\textwidth]{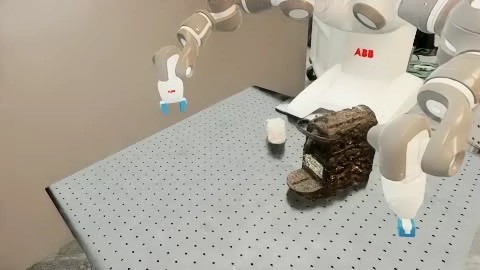} &
        \includegraphics[width=0.30\textwidth]{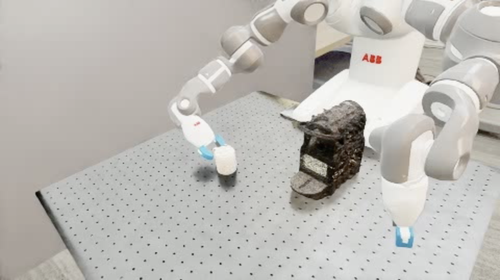} \\
    \end{tabular}
    \caption{\textbf{Real2Render2Real} (Views From Both Cameras Shown). Base augmentations used in the main R2R2R experiments include: random sphere lighting, camera pose perturbation, robot initial joint perturbation, and randomized object initialization uniformly distributed via manual parameters.}
    \label{fig:base_abl}
\end{figure}

\begin{figure}[h!]
    \centering
    \begin{tabular}{ccc}
        \rotatebox[origin=c]{180}{\includegraphics[width=0.30\textwidth]{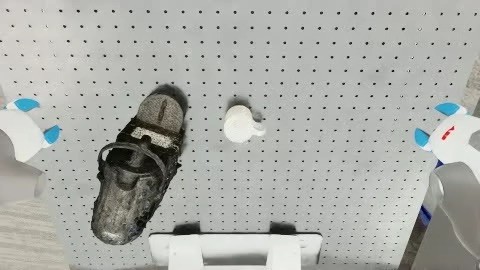}} &
        \rotatebox[origin=c]{180}{\includegraphics[width=0.30\textwidth]{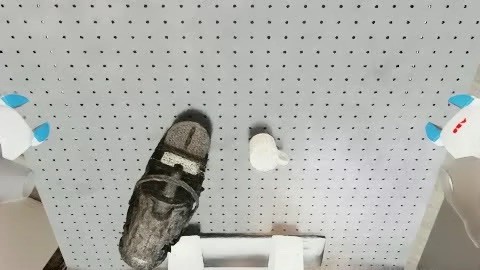}} &
        \rotatebox[origin=c]{180}{\includegraphics[width=0.30\textwidth]{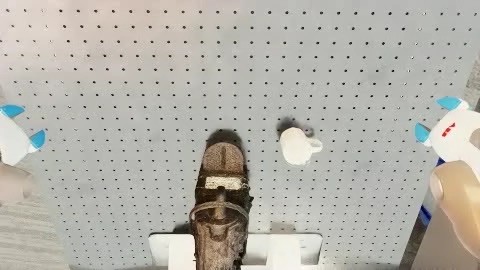}} \\
        \rotatebox[origin=c]{180}{\includegraphics[width=0.30\textwidth]{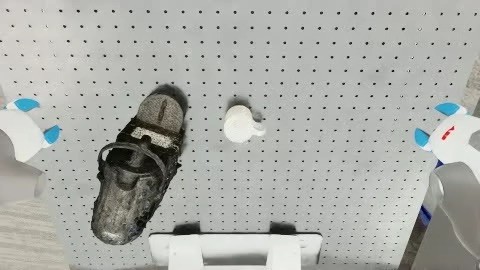}} &
        \rotatebox[origin=c]{180}{\includegraphics[width=0.30\textwidth]{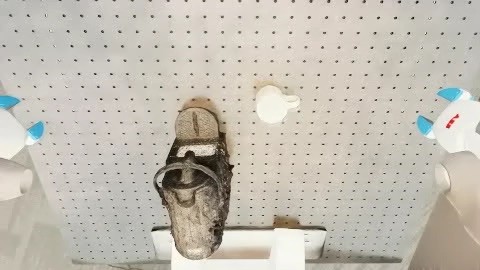}} &
        \rotatebox[origin=c]{180}{\includegraphics[width=0.30\textwidth]{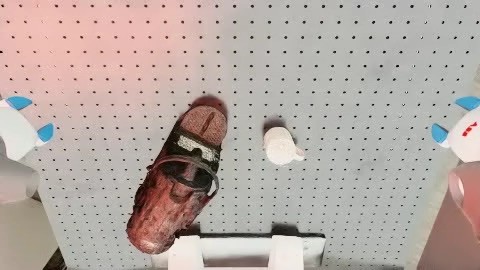}} \\
    \end{tabular}
    \caption{\textbf{Trajectory Interpolation Turned Off} (Top Camera Views Shown). \textit{Note the fixed configuration of the mug with respect to the coffee maker.} With trajectory interpolation off for multiple rigid bodies, we may only densely follow the tracked trajectories shown in the video demonstration. Without it, the only method for increasing trajectory diversity would be to augment with part trajectories from adding/tracking additional demonstration videos.}
    \label{fig:base_abl}
\end{figure}

\begin{figure}[h!]
    \centering
    \begin{tabular}{ccc}
        \rotatebox[origin=c]{180}{\includegraphics[width=0.30\textwidth]{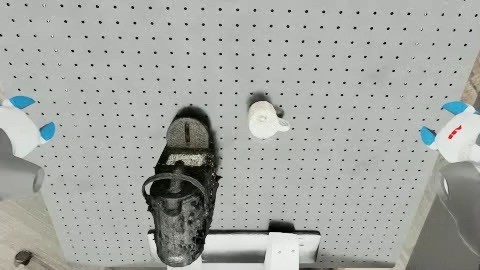}} &
        \rotatebox[origin=c]{180}{\includegraphics[width=0.30\textwidth]{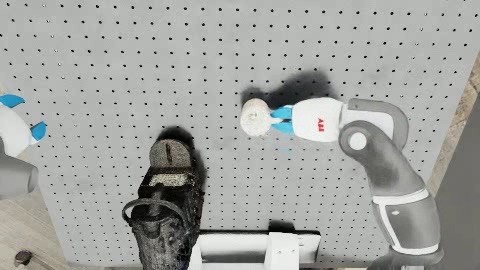}} &
        \rotatebox[origin=c]{180}{\includegraphics[width=0.30\textwidth]{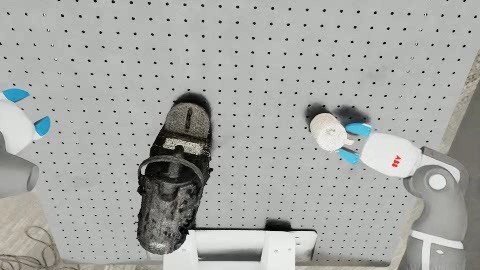}} \\
        \rotatebox[origin=c]{180}{\includegraphics[width=0.30\textwidth]{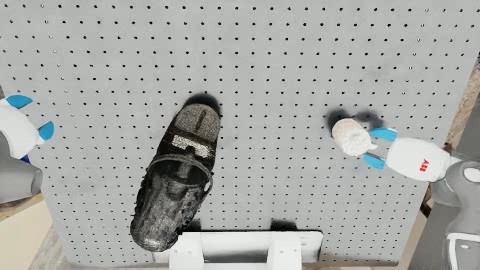}} &
        \rotatebox[origin=c]{180}{\includegraphics[width=0.30\textwidth]{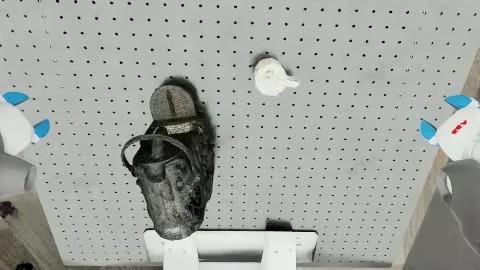}} &
        \rotatebox[origin=c]{180}{\includegraphics[width=0.30\textwidth]{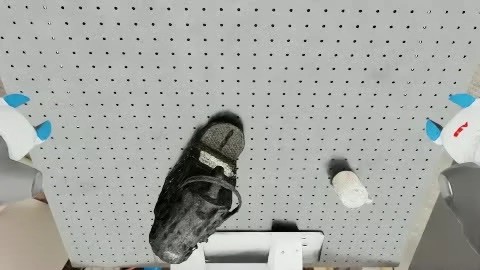}} \\
    \end{tabular}
    \caption{\textbf{Random Lighting Augmentation Turned Off} (Top Camera Views Shown). We turn off the randomized sphere light sources with varying colors and intensities. Uniform lighting is available from the only light source in the render scene -- the skybox/dome-light asset.}
    \label{fig:lighting_abl}
\end{figure}
\newpage

\subsection{Upfront Processing Time Until Generation}

\begin{table}[h]
\small
\centering
\begin{tabular}{lc}
\toprule
Task & Time to Complete\\
\midrule
Scanning & 1 min \\
Demonstration & $<$10 sec \\
GARField Segmentation~\cite{garfield2024} & 2 min \\
3DGS Optimization & 1 min \\
4D-DPM Tracking~\cite{kerr2024robot} & 3 mins \\
SuGaR~\cite{guedon2023sugar} Meshification & 2 mins \\
Asset into IsaacLab & 1 min \\
\bottomrule
\end{tabular}
\vspace{4pt}
\caption{\textbf{Upfront Processing Time per Task Prior to R2R2R Data Generation.}  
Breakdown of one-time preprocessing steps required to convert a demonstration video and scanned asset into a renderer-ready format for R2R2R. These steps include segmentation, tracking, meshification, and asset import.}
\label{tab:task_upfront}
\vspace{-1.5em}
\end{table}

\subsection{Data Collection and Generation Time}

\begin{table}[h]
\small
\centering
\begin{tabular}{lcc}
\toprule
Task & Teleop, 150 demos, 1 operator & \algabbr, 1k demos, 1 GPU \\
\midrule
Pick up the tiger & 60 mins & 26.15 mins \\
Put the mug on the coffee maker & 86 mins &  38.22 mins \\
Pick up the package with both hands & 90 mins & 13.97 mins \\
Open the drawer & 71 mins & 16.95 mins \\
Turn the faucet off & 104 mins & 16.67 mins \\
\bottomrule
\end{tabular}
\vspace{4pt}
\caption{Time taken per task to either collect 150 demos through teleoperation with one human operator or to generate 1000 synthetic demos with \algabbr. Note: \algabbr generation times do not include the upfront processing time until generation.}
\label{tab:task_data_coll_time}
\vspace{-1.5em}
\end{table}
\newpage

\subsection{Extended Training Details}
\label{sec:appendix_extended_training}
We provide hyperparameters for training diffusion policy~\cite{chi2023diffusion} from scratch and fine-tuning $\pi_0$-FAST~\cite{pertsch2025fast} with LoRA in \cref{tbl:diffusion_hyper} and \cref{tbl:pi0_hyper}. 

\begin{table*}[h!]
    \centering
    \begin{tabular}{cc}
    \toprule
    Config                 & Value
    \\ \hline
    optimizer              & AdamW~\citep{loshchilov2017decoupled}                                                                             \\
    base learning rate     & 2e-4                                                                            \\
    learning rate schedule & cosine decay~\citep{loshchilov2016sgdr}                                                                      \\
    batch size             & 64                                                                              \\
    weight decay           & 0.09                                                                              \\
    optimizer momentum     & $\beta_1, \beta_2$ = 0.9, 0.999~\citep{chen2020generative}                                                      \\
    warm up steps~\citep{Goyal2017b}          & 500                                                                            \\
    
    total steps           & 100,000                                                                          \\
    observation history           & 4                                                                          \\
    action dimension           & 20 (YuMi)                                                                           \\
    proprio format           & absolute eef xyz, 6d rotation, absolute gripper position                                                                          \\
    action format           & delta eef xyz, 6d rotation, absolute gripper position                                                                           \\
    action horizon           & 16                                                                          \\
    observation resolution           & 448                                                                          \\
    \bottomrule
    \end{tabular}
    \caption{Diffusion Policy Hyperparameters}
    \label{tbl:diffusion_hyper}
\end{table*}

\begin{table*}[h!]
    \centering
    \begin{tabular}{cc}
    \toprule
    Config                 & Value    
        \vspace{8.0px}
    \\ \hline

    optimizer              & AdamW~\citep{loshchilov2017decoupled}                                                                             \\
    base learning rate     & 2.5e-5                                                                            \\
    learning rate schedule & cosine decay~\citep{loshchilov2016sgdr}                                                                      \\
    batch size             & 32                                                                              \\
    weight decay           & 0.09                                                                              \\
    LoRA Rank           & 16                                                                          \\
    LoRA Alpha           & 16                                                                          \\
    optimizer momentum     & $\beta_1, \beta_2$ = 0.9, 0.95~\citep{chen2020generative}                                                      \\
    warm up steps~\citep{Goyal2017b}          & 1000                                                                            \\
    total steps           & 30,000                                                                          \\
    action/proprio dimension           & 16 (YuMi) 8 (Franka)                                                                          \\
    proprio format           & absolute joints positions, absolute gripper position                                                                          \\
    action format           & delta joints, absolute gripper position                                                                          \\
    action horizon           & 10                                                                          \\
    observation resolution           & 224                                                                          \\
    \bottomrule
    \end{tabular}
    \caption{$\pi_0$-FAST hyperparameters}
    \label{tbl:pi0_hyper}
\end{table*}
\newpage 

\subsection{Statistical Comparison Between Teleoperation and R2R2R Data Efficacy}
\label{sec:appendix_p_test}
To evaluate whether R2R2R-generated data yields performance comparable to human teleoperation, we apply the Two One-Sided Tests (TOST) procedure across all tasks and policies. Unlike traditional significance tests that ask whether two conditions differ, TOST tests whether the difference between them is small enough to be considered practically negligible. Specifically, we test whether the absolute difference in success rates falls within a $\pm5\%$ margin—chosen to reflect a practically insignificant difference for robot policy success rates.

As shown in \cref{tab:tost_results}, no individual task satisfies both conditions required for statistical equivalence (i.e., both p-values below 0.05). However, the results consistently show no strong evidence that either R2R2R or teleoperation outperforms the other. In particular, the global test across all tasks yields one p-value below 0.05 and one above, suggesting performance is similar but not provably equivalent under the chosen threshold. These results support the interpretation that R2R2R can match the effectiveness of teleoperation across the evaluated tasks, while offering a significantly more scalable method for data generation.

\begin{table}[htbp]
    \centering
    \begin{tabular}{l l cc}
        \toprule
        \textbf{Task} & \textbf{Policy} & \textbf{TOST lower p} & \textbf{TOST upper p} \\
        \midrule
        \multirow{2}{*}{Pick up the toy tiger} 
            & Diffusion Policy & 0.2656 & 0.5359 \\
            & $\pi_0$-FAST (Finetuned) & 0.6891 & 0.1429 \\
        \midrule
        \multirow{2}{*}{Put the mug on the coffee maker} 
            & Diffusion Policy & 0.5349 & 0.2712 \\
            & $\pi_0$-FAST (Finetuned) & 0.6891 & 0.1429 \\
        \midrule
        \multirow{2}{*}{Turn the faucet off} 
            & Diffusion Policy & 0.6891 & 0.1429 \\
            & $\pi_0$-FAST (Finetuned) & 0.3729 & 0.3729 \\
        \midrule
        \multirow{2}{*}{Open the Drawer} 
            & Diffusion Policy & 0.2656 & 0.5359 \\
            & $\pi_0$-FAST (Finetuned) & 0.8051 & 0.0806 \\
        \midrule
        \multirow{2}{*}{Pick up the package with both hands} 
            & Diffusion Policy & 0.1429 & 0.6891 \\
            & $\pi_0$-FAST (Finetuned) & 0.3955 & 0.3955 \\
        \midrule
        Overall (All Tasks) & -- & 0.4271 & 0.0497 \\
        \bottomrule
    \end{tabular}
    \vspace{4pt}
    \caption{
        \textbf{Equivalence testing (TOST) between human teleoperation (150 trajectories) and R2R2R-generated data (1,000 trajectories).}
        We report the p-values from Two One-Sided Tests (TOST) applied to each task and policy, using a $\pm5\%$ success rate margin as the equivalence threshold.
        The ``lower p" tests whether R2R2R performs \emph{no worse} than teleoperation by more than 5\%, while the ``upper p" tests whether teleoperation performs \emph{no worse} than R2R2R.
        Statistical equivalence is only confirmed when \emph{both} p-values fall below 0.05.
    }
    
    \label{tab:tost_results}
\end{table}

\newpage

\bibliography{main}  

\end{document}